\pdfoutput=1

\documentclass[11pt]{article}

\usepackage{acl}

\usepackage{times}
\usepackage{latexsym}
\usepackage{slashbox}

\usepackage[T1]{fontenc}

\usepackage[utf8]{inputenc}

\usepackage{microtype}
\usepackage{multirow}
\usepackage{wrapfig}

\usepackage{hyperref}       
\usepackage{url}            
\usepackage{booktabs}       
\usepackage{amsfonts}       
\usepackage{nicefrac}       
\usepackage{microtype}      
\usepackage{xcolor}         
\usepackage{graphicx}
\usepackage{rotating}
\usepackage{mdframed}

\definecolor{MidnightBlue}{RGB}{0,51,102}

\hypersetup{
    colorlinks=true,
    linkcolor=MidnightBlue,
    citecolor=black
}
\usepackage{cleveref}

\newcommand{\craftico}[0]{\includegraphics[width=.05\textwidth]{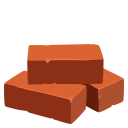}}

\title{\craftico~CRAFT: A Benchmark for Causal Reasoning About\\ Forces and inTeractions}

\author{
Tayfun Ates\textsuperscript{1,}\thanks{~~indicates equal contributions.}\quad
M. Samil Atesoglu\textsuperscript{1,}$^*$\quad
Cagatay Yigit\textsuperscript{1,}$^*$\quad
Ilker Kesen\textsuperscript{2,3} \\ \quad
{\bf Mert Kobas\textsuperscript{4}}\quad
{\bf Erkut Erdem\textsuperscript{1,2}}\quad
{\bf Aykut Erdem\textsuperscript{2,3}}\quad
{\bf Tilbe Goksun\textsuperscript{4}}\quad
{\bf Deniz Yuret\textsuperscript{2,3}}
\\

\textsuperscript{1} Hacettepe University, Computer Engineering Department~ \textsuperscript{2} Ko\c{c} University Is Bank AI Center \\
\textsuperscript{3} Ko\c{c} University, Computer Engineering Department~\textsuperscript{4} Ko\c{c} University, Psychology Department\\
}

\begin{document}

\maketitle

\begin{abstract}
Humans are able to perceive, understand and reason about causal events. Developing models with similar physical and causal understanding capabilities is a long-standing goal of artificial intelligence. As a step towards this direction, we introduce CRAFT\footnote{Data and code available on our project website at \href{https://sites.google.com/view/craft-benchmark}{https://sites.google.com/view/craft-benchmark}}, a new video question answering dataset that requires causal reasoning about physical forces and object interactions. It contains 58K video and question pairs that are generated from 10K videos from 20 different virtual environments, containing various objects in motion that interact with each other and the scene. Two question categories in CRAFT include previously studied \emph{descriptive} and \emph{counterfactual} questions. Additionally, inspired by the Force Dynamics Theory in cognitive linguistics, we introduce a new \emph{causal} question category that involves understanding the causal interactions between objects through notions like \emph{cause}, \emph{enable}, and \emph{prevent}. Our results show that even though the questions in CRAFT are easy for humans, the tested baseline models, including existing state-of-the-art methods, do not yet deal with the challenges posed in our benchmark.
\end{abstract}

\section{Introduction}
Causal reasoning is a key cognitive capability that involves making predictions about physical objects and their interactions. Cognitive scientists have mainly studied causal reasoning as simple causes or chains of events~\cite{michotte1963perception,baillargeon1994physical,saxe2005secret}, rather than processing of complex causal scenes, see~\cite{goksun2013forces,george2019any}. Referring to the interactions of multiple forces, the Force Dynamics Theory emphasizes the processing and reasoning of complex scenes, and how causal language defines the patterns of forces in causal events~\cite{wolff2007representing}.

In the past decade, though artificial learning systems have shown astonishing progress in natural language and image understanding, there are some tasks in which these systems are still significantly below human performance. One such challenging research area includes reasoning about physical actions of objects in complex causal scenes. In this paper, we explore how language and vision interact with each other in making plausible projections about causal reasoning, and analyze how well the existing neural models understand and reason about physical and causal relationships between dynamic objects in a scene through images and~text.

\begin{figure*}[!t]
\centering
\includegraphics[width=\textwidth]{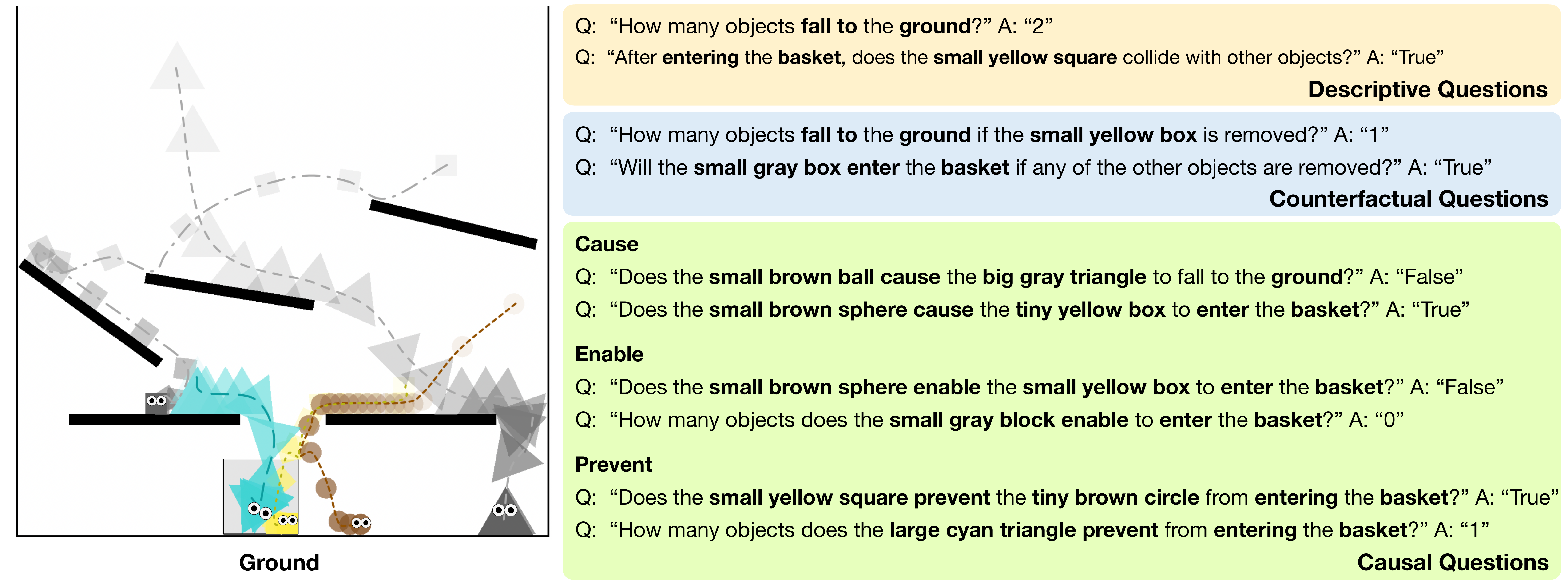}
\caption{\textbf{Example CRAFT questions generated for a sample scene.} There are 48 different tasks divided into three distinct categories for 20 different scenes. Besides having tasks questioning descriptive properties, possibly needing temporal reasoning, CRAFT introduces challenges including more complex tasks requiring single or multiple counterfactual analysis or understanding object intentions for deep causal reasoning.}
\label{svqa_eg1}
\end{figure*}

We propose a new video question answering dataset, named CRAFT (Causal Reasoning About Forces and inTeractions), which is designed to be complex for artificial models and simple for humans. Our dataset contains synthetically generated videos of 2D scenes with accompanying questions. Its most prominent features are that it contains video clips with complex physical interactions between objects, and questions that test strong reasoning capabilities. Answering our \emph{causal} questions needs comprehending what is being asked, identifying objects in the scene, tracking their states in relation to other objects, which in turn can be attributed to different semantic categories of causes (\emph{cause, enable} or \emph{prevent}) that highlight unique patterns of causal forces in events -- in line with the Force Dynamics Theory. In CRAFT, there are also some \emph{descriptive} and \emph{counterfactual} questions, the latter requiring understanding what would have happened after an intervention, i.e. a slight change in the scene~\citep{wolff2013}. Figure~\ref{svqa_eg1} shows sample questions from different question types, which are explained in detail in the subsequent sections.

\section{Related Work}
\textbf{Visual Question Answering.} Existing visual question answering (VQA) datasets can be categorized along two dimensions. The first dimension is the type of visual data, which includes either real world images~\citep{malinowski2014multi, ren2015exploring, antol2015vqa, zhu2016visual7w, goyal2017making} or videos~\citep{tapaswi2016movieqa, lei2018tvqa}, or synthetically created content~\citep{johnson2017clevr, zhang2016yin, yi2019clevrer}. The second is at how the questions and answers are  collected, which are usually done via crowdsourcing~\citep{malinowski2014multi, antol2015vqa} or by automatic means~\citep{ren2015exploring, lin2014microsoft, johnson2017clevr}. A key challenge for creating a good VQA dataset lies in minimizing the dataset bias. A model may exploit such biases and cheat the task by learning some shortcuts. In our work, we generate questions about simulated scenes using a pre-defined set of templates by considering some heuristics to eliminate strong biases. Compared to the existing VQA datasets, CRAFT is specifically designed to test models' understanding of dynamic state changes of the objects in a scene. Although some prior work focuses on temporal reasoning~\citep{lei2018tvqa, yu2019activitynet, lei2019tvqa+,girdhar2020}, they do not require the models to have a deep understanding of physics and/or imagine the consequence of certain actions to answer the questions, the only exceptions being TIWIQ~\citep{wagner2018answering}, CLEVRER~\citep{yi2019clevrer}, CLEVR\_HYP~\citep{sampat2021clevrhyp} and TVR~\citep{Hong_2021_CVPR} datasets. In these datasets, there exist hypothetical questions that require mental simulations about the consequences of performing certain actions or the lack of specific actions or objects. These datasets have received interest in developing neuro-symbolic reasoning models with physical understanding capabilities \citep{ding2020,chen2021,ding2021dynamic}. CRAFT shares a similar design goal with the aforementioned datasets -- but the scenes in our benchmark are temporally more complex.

\noindent\textbf{Causal Reasoning in Cognitive Science.} 
Different theories have been proposed by cognitive scientists to model how humans learn, experience, and reason about causal events, Mental Model Theory \citep{khemlani2014causal}, Causal Model Theory \citep{sloman2009causal}, and Force Dynamics Theory \citep{wolff2015causal} to name a few. Among these, building upon the work of Talmy (1988), the Force Dynamics Theory represents a variety of causal relationships such as cause, enable, and prevent between two main entities, an affector and a patient (i.e. the object the affector acts on). The theory emphasizes that causative verbs map onto these different spatial arrays of forces within complex causal scenes. Studies with speakers of different languages such as English, Russian, and German suggest that adults distinctly represent these semantic event categories~\cite{wolff2003models,wolff2005expressing}. Similarly, \mbox{5- to 6-year-old} children perceive the interactions of forces underlying the semantic categories of cause, enable, and prevent~\cite{goksun2013forces} and make inferences about these events~\cite{george2019any}. To our knowledge, our work is the first attempt at integrating these complex relationships in a VQA setup to test causal reasoning capabilities of machines.\vspace{1em}

\noindent\textbf{Understanding Physics in Artificial Intelligence.} 
Lately, there has been a growing interest within the community in developing datasets and models to evaluate the ability of understanding and reasoning about the physical world. A notable amount of these efforts focuses on physical scene understanding. For instance, some researchers have explored the problem of predicting whether a set of objects are in stable configuration or not \citep{mottaghi2016newtonian} or if not where they fall \citep{lerer2016learning}. Others have tried to estimate a motion trajectory of a query object under different forces \citep{mottaghi2016newtonian} or developed methods to build a stack configuration of the objects from scratch through a planning algorithm \citep{janner2019reasoning}. \citet{li2018learning} suggested to represent rigid bodies, fluids, and deformable objects as a collection of particles and used this representation to learn how to manipulate them. Recently, \citet{bakhtin2019phyre} and \citet{allen2020tools} created the PHYRE and the Tools benchmarks, respectively, which both include different types of 2D environments. An agent must reason about the scene and predict the outcomes of possible actions in order to solve the task associated with the environment. CoPhy~\citep{bradel2020} is another recent work, which deals with physical reasoning prediction about counterfactual interventions. Although these works involve complicated physical reasoning tasks, the language component is largely missing. As mentioned, \citet{wagner2018answering}, \citet{yi2019clevrer} and \citet{sampat2021clevrhyp} created VQA datasets for intuitive physics, but they lack visual variations unlike PHYRE and Tools. Though less studied, there are also some efforts in the NLP community to evaluate physical reasoning abilities of language models. \citet{bisk2020piqa} proposed the PIQA dataset that involves a binary choice task about daily activities regarding physical commonsense. Similarly, \citet{aroca2021prost} presented the PROST benchmark which includes questions that are designed to probe
language models in a zero-shot setting and focuses on concepts like gravitational forces, physical attributes and object affordances.

Our CRAFT dataset aims to combine the best of both worlds. In addition to the two types of questions investigated in CLEVRER~\citep{yi2019clevrer}, namely \emph{descriptive} and \emph{counterfactual}, CRAFT also includes questions that need reasoning about \emph{causal} interactions through the concepts like \emph{cause}, \emph{enable}, and \emph{prevent}. To succeed in these tasks, models need to learn the semantics of each verb category that specifies different kinds of object interactions and their outcomes, i.e. to gain an understanding of a kind of commonsense knowledge.

\section{The CRAFT Dataset}
\label{sec:craft-dataset}
CRAFT is built to evaluate temporal and causal reasoning capabilities of existing algorithms on video clips of 2D simulations and related questions. The dataset has approximately 57K question and video pairs, which are created from 10K videos. It is split into train, validation, and test sets with a 60:20:20 ratio per video basis, meaning that video clips in the training set are not seen in the validation or test set. Moreover, we have two different settings, an \emph{easy setting} and a \emph{hard setting}. They differ from each other in the way how the test split is chosen. In the hard setting, we deliberately use scene layouts that are not seen during training in picking the video and question pairs. The easy setting does not have this constraint. In the easy setting, there are 35K, 12K, and 11K question and video pairs in the train, validation and test splits, whereas in the hard setting these numbers are 35K, 11K and 12K, respectively. We provide an example set of questions from CRAFT in Figure \ref{svqa_eg1}. \vspace{0.25em} 

\begin{figure*}[!t]
\centering
\includegraphics[width=0.925\linewidth]{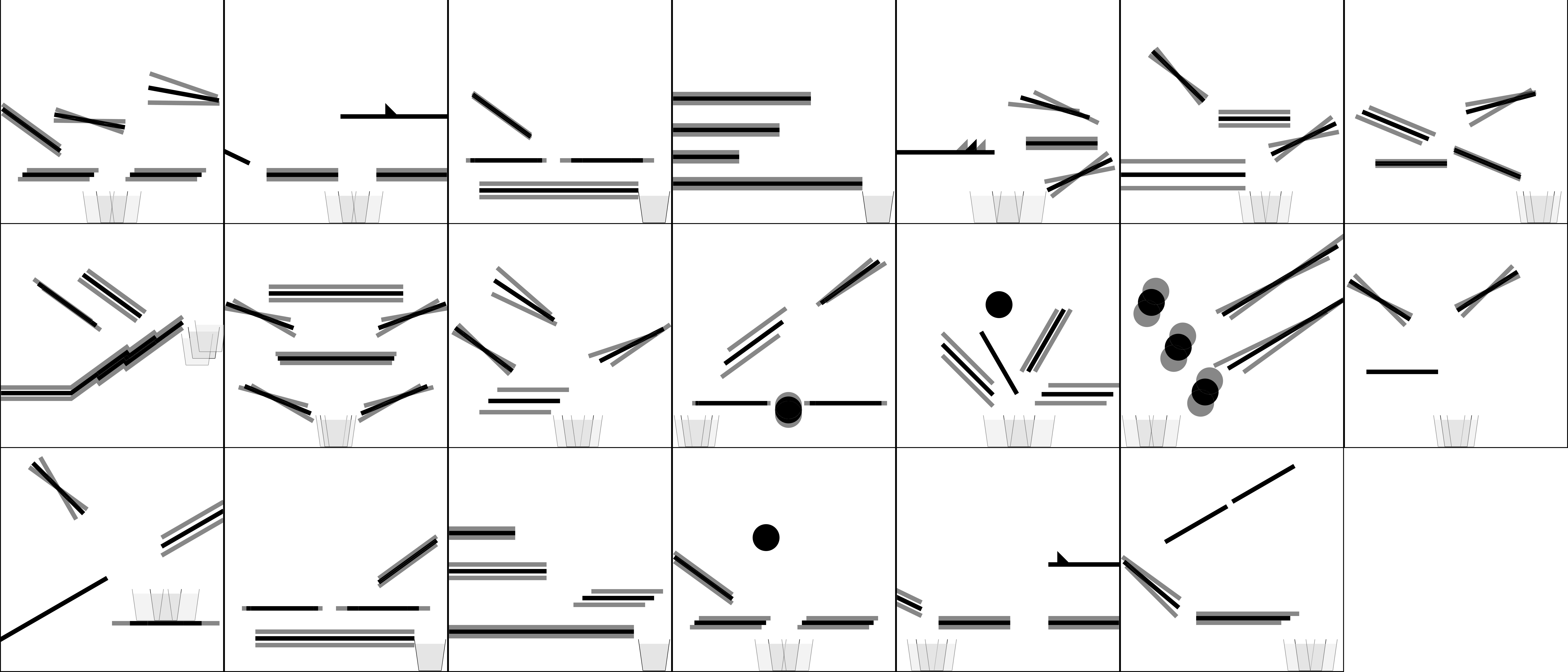}
\caption{\textbf{Random configurations of static scene element properties for each scene.} The opaque regions show the mean value for that element, whereas the overlayed regions show the extreme values. Although these changes may seem subtle, they provide a wide variety in terms of scene dynamics.}
\label{static_objects}
\end{figure*}

\noindent\textbf{Video Generation.} We use Box2D physics simulator \citep{catto2010} to create our virtual scenes. There are 20 distinct scene layouts from which 10 seconds of video clips are collected with a spatial resolution of $256\times 256$ pixels. Besides generating original simulation video, CRAFT scripts also generate variation videos by removing each object of the same video from the scene. These variation videos help question generation script to provide answer for certain types of questions, as explained~later.\vspace{0.25em}

\noindent\textbf{Objects.} Each scene is composed of both \emph{static scene elements} and \emph{dynamic objects}, containing variable number of and different type of these elements and objects. There are 7 static scene elements (\textit{ramp}, \textit{platform}, \textit{button}, \textit{basket}, \textit{left wall}, \textit{right wall}, \textit{ground}). These elements are all drawn in \textbf{black} color in order to differentiate them from the dynamic objects. Their attributes such as position or orientation are decided at the beginning of a simulation and then they are kept fixed throughout the video sequence. The values of these attributes are assigned randomly from sets of different intervals which are predefined for each type of scene as in Figure \ref{static_objects}. The set of the dynamic objects contains 3 shapes (\textit{cube}, \textit{triangle}, \textit{circle}), 2 sizes (\textit{small}, \textit{large}), and 8 colors (\textit{gray}, \textit{red}, \textit{blue}, \textit{green}, \textit{brown}, \textit{purple}, \textit{cyan}, \textit{yellow}). Attributes of dynamic objects, in contrast, are in continuous change throughout the sequence due to the gravity or the interactions that they are subject to, until they rest. \vspace{0.25em}

\begin{figure}[!t]
\centering
\includegraphics[width=\linewidth]{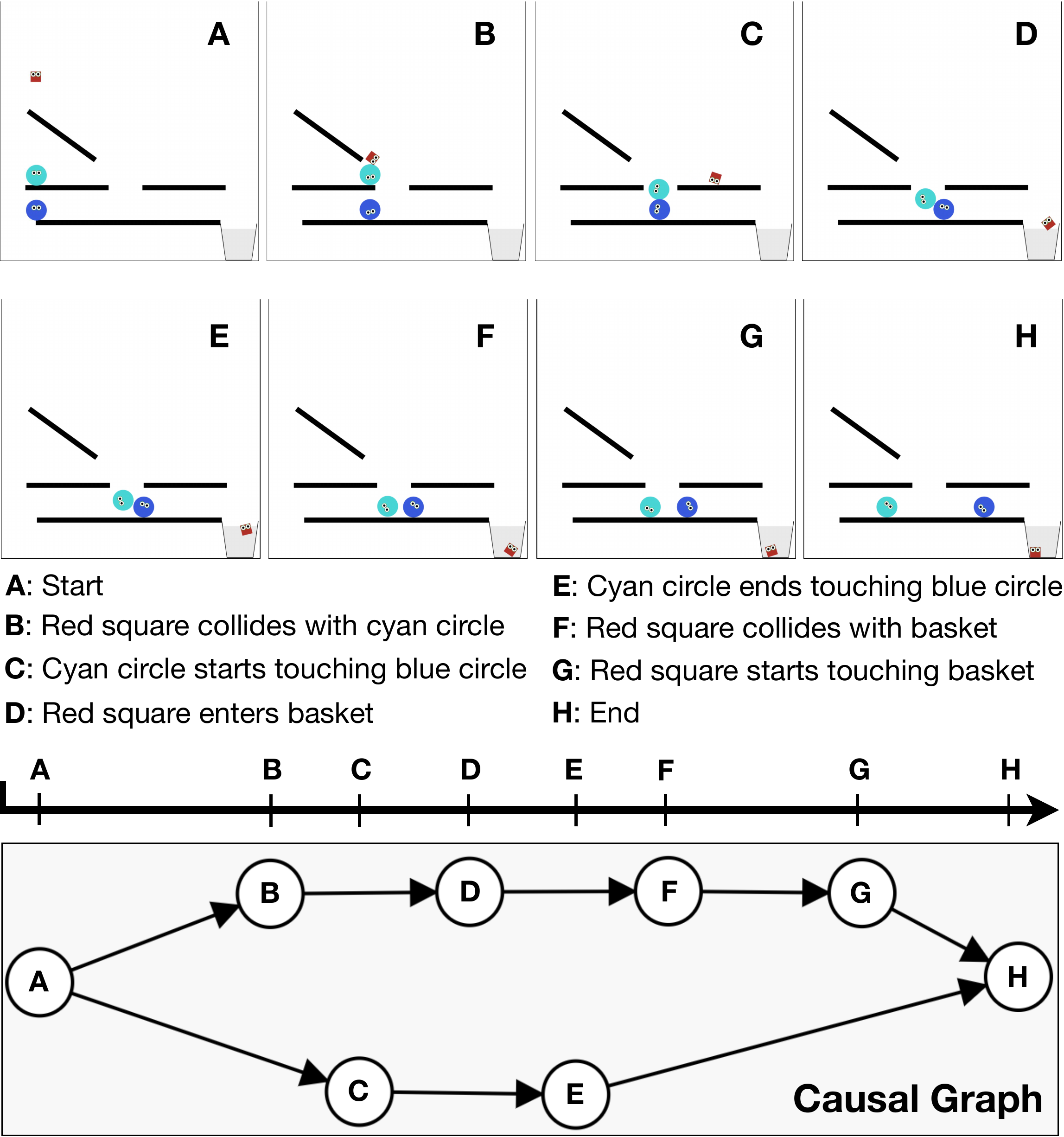}
\caption{\textbf{A simple causal graph.} The causal graph is a graphical summary of the events that occur in a simulation. For the sake of simplicity, here we only include the interactions between the dynamic objects and the basket, and moreover, the scene is uncomplicated that there is no intermediate branching in the causal graph.}
\label{casualgraph}
\end{figure}

\noindent\textbf{Events.} To formally represent the dynamical interactions in the simulations, we extract different types of events: \textit{Start}, \textit{End}, \textit{Collision}, \textit{Touch Start}, \textit{Touch End}, and \textit{Enter Basket}. \textit{Start} and \textit{End} events represent the start and the end of the simulations, respectively. Although we mainly question \textit{Collision} events in our tasks, we want models to understand the difference between a collision and rolling on a ramp or a platform or two objects moving together. Therefore, we also extract \textit{Touch Start}, \textit{Touch End} events. Finally, \textit{Enter Basket} event is triggered if the object enters the basket in the scene. All events happening a simulation are represented as a causal graph, which is also key for the question generator to extract causal relationships in an easy manner. Causal graph is a directed graph where events are represented as nodes. Each edge represents a cause relation where the source event is considered as the cause of target event because of the shared objects between them. We demonstrate the causal graph of a sample simulation in Figure~\ref{casualgraph}.\vspace{0.25em} 

\noindent\textbf{Simulation Representation.} A simulation instance is represented by three different data structures, \textit{the initial state of the scene}, \textit{the final state of the scene}, and \textit{the causal graph of extracted events}. The initial and final state of a scene refers to the information regarding the objects' static and dynamic attributes such as color, position, shape, and velocity at the start or at the end of the simulation, respectively. The final state is important as it bears causal relationships between the events of a simulation. Together these information sources have sufficient information to find the correct answers to CRAFT questions. Our simulation system also allows us to generate scene graphs like the ones used in CLEVR \citep{johnson2017clevr}, though we have not investigated it yet.

\noindent\textbf{Question Generation.} Each CRAFT question is expressed with a functional program as in CLEVR. We use a different set of functional modules for our programs extending the CLEVR approach. For example, our module set includes, but not limited to, functions which can filter events such as \textit{Enter Basket} and \textit{Collision}, and functions which can filter objects based on whether they are stationary at the start or the end of the video. List of our functional modules and some example programs are provided in Appendices~\ref{app_funcmodules} and \ref{app_programs} in the supplementary material, respectively. Moreover, we use different sets of word synonyms and allow question text to be paraphrased for language variety similar to CLEVR. Our preliminary analysis revealed that human performances in some questions were poor. When investigated, we figured out that these questions seem to be counter-intuitive to humans. Humans do not accurately reason about the objects for some counterfactual cases as subtle changes in the scenes result in very different outcomes. Hence, in finalizing our dataset, we applied minor random perturbations to each dynamic object in a video to verify whether the same answer can be obtained for all such cases, and excluded those questions that did not pass this verification step.\vspace{0.25em}

\noindent\textbf{Question Types.} CRAFT has 48 different question types under 3 different categories, namely \textit{Causal, Descriptive, Counterfactual}. Among these, \textit{Descriptive} questions mainly require extracting the attributes of objects, but some of them, especially those involving counting, need temporal analysis as well. Our dataset extends CLEVRER by \citet{yi2019clevrer} with different types of events and multiple environments. \textit{Counterfactual} questions require understanding what would happen if one of the objects was removed from the scene. Exclusive to CRAFT, some \textit{Counterfactual} questions (\textit{``Will the small gray circle enter the basket if any of the other objects are removed?''}) require multiple counterfactual simulations to be explored. As an extension to \textit{Counterfactual} questions, \textit{Causal} questions require grasping what is happening inside both the original video and the counterfactual video. In other words, models must infer whether an object is causing or enabling an event or preventing it by comparing the input video and the counterfactual video that should be simulated somehow. In the question text, the affector and the patient objects are explicitly specified. Some questions even include multiple patients. In particular, distinct causative verbs are mapped onto these three classes of causal events (Table~\ref{tab:causal_verbs}).\vspace{0.25em} 

\begin{table}[!t]
\caption{The list of causative verbs and their categories which are considered in CRAFT.}
\centering
\resizebox{0.9\linewidth}{!}{
\begin{tabular}{ll}
\toprule
\bf{Category} & \bf{Verbs} \\
\midrule
Prevent & \textit{prevent, keep, hold, block, hinder}\\
Enable & \textit{enable, help, allow}\\
Cause & \textit{cause, stimulate, trigger}\\
\bottomrule
\end{tabular}}
\label{tab:causal_verbs}
\end{table}

In order to have a better understanding of the differences between \textit{Enable}, \textit{Cause}, and \textit{Prevent} questions, one should understand the \emph{intention} of the objects. We identify the intention in a simulation by examining the initial velocity of the corresponding object. Inspired by the recent findings in cognitive linguistics~\cite{beller2020language}, we take having a velocity as an indication of an intention. In that regard, an affector can only enable a patient to complete the task if the patient is originally intended to do it but fails without the affector. Similarly, an affector can only cause a patient to do the task if the patient is not intended to execute it. Moreover, an affector can only prevent a patient from completing the task if the patient is intended to do it and succeeds without the affector.\vspace{0.25em}

\noindent\textbf{Variations in Natural Language.} In datasets that involve a natural language component, it is crucial to have language variety. To improve this property, CRAFT data generation scripts for questions, first allow multiple paraphrased versions of the same text to be generated to represent the same task. For a question sample, a paraphrased version of the corresponding task is chosen randomly by filling the object templates. Second, CRAFT enables synonyms of certain words to be integrated. We choose a base word and create its synonyms inside the CRAFT context. Similar to question paraphrases, the base word is replaced by a synonym randomly at run-time. All synonyms including the base word have equal chance to be included in the question text. This is handled by word suffixes and verb conjugations by preserving English grammar.\vspace{0.25em} 

\begin{figure}[!t]
\centering
\includegraphics[scale=0.51]{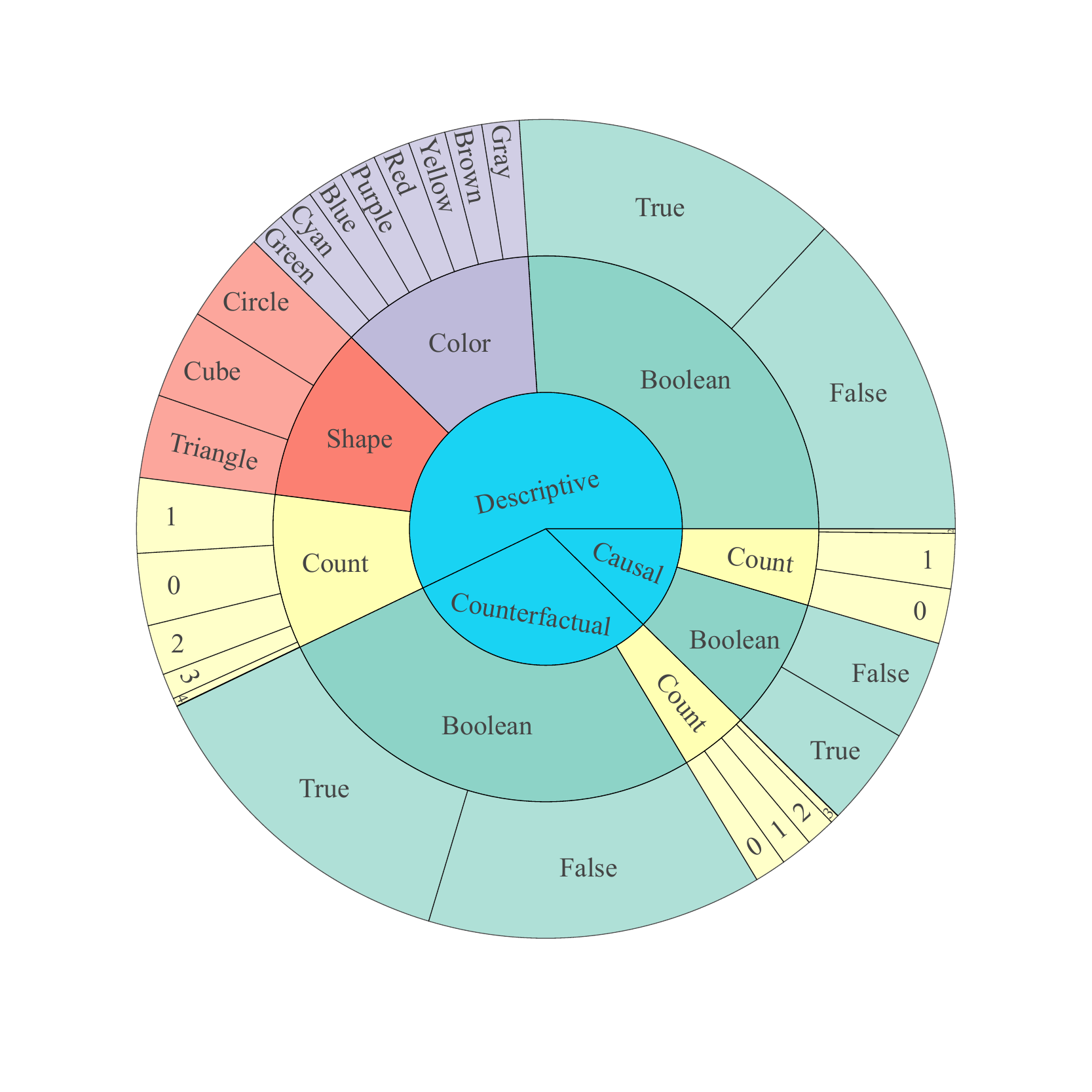}
\caption{\textbf{Distribution of question types and answers in CRAFT.} Innermost layer represents the distribution of the questions  for different task categories. Middle layer illustrates the distribution of the answer types for each task category. Outermost layer represents the distribution of answers for each answer type.}
\label{stats}
\end{figure}

\noindent\textbf{Bias Reduction.} CRAFT contains simulations from different scenes to increase the variety in the visual domain. This makes reducing the dataset biases difficult because of the multiplicity in the number of the domains (textual and visual). Our data generation process enforces different simulation and task pairs to have uniform answer distributions while trying to keep overall answer distribution as uniform as possible. Our aim is to make it harder for the models to find simple shortcuts by predicting the task identifier, the simulation identifier, or both, instead of understanding the scene dynamics and the question. Figure~\ref{stats} shows the answer distributions for the question categories in CRAFT. 

\section{Experimental Analysis}
In this section, we evaluate the performances of a wide range of baseline models on CRAFT. We also analyze how these performances relate with that of humans in understanding physical interactions between the objects and the environment.

\subsection{Baselines}
In our experiments, we consider several weak and strong baselines including some state-of-the-art visual reasoning approaches. 

\noindent \textbf{Heuristic} models either perform random guesses or follow simple rules. \textbf{Random} model uniformly samples a random answer from the full answer space, whereas \textbf{Answer Type Based Random} model (AT-Random) makes random guesses based on the answer type (e.g. color, shape, boolean). \textbf{Most Frequent Answer} baseline (MFA) employs a simple heuristics and answers all the questions by using the most frequent answer in the training split. \textbf{Answer Type based Most Frequent Answer} model (AT-MFA) performs the same heuristics by taking the answer types into account similar to AT-Random baseline. 

\noindent \textbf{Text-only models} ignore simulations, and do not use any visual information related to input simulations. \textbf{LSTM} model is another image-blind baseline that processes the question with an LSTM \citep{hochreiter1997long}, and then predicts an answer to a given question ignoring the visual input. In addition to the LSTM baseline, we experimented with \textbf{BERT} \cite{devlin-etal-2019-bert} by using the \texttt{CLS} token embedding as question representation to predict answers.

\noindent \textbf{LSTM-CNN} baseline integrates both visual and textual cues by extending the LSTM model to additionally consider the features extracted from the 
a pretrained ResNet-18 model. We evaluate both (non-temporal) single frame and video versions. In the former, each video is encoded by taking into account either the first frame or the last frame, which are referred to as \textbf{LSTM-CNN-F} and \textbf{LSTM-CNN-L}, respectively. The video version, which we call \textbf{LSTM-CNN-V}, processes downsampled videos by using R3D \citep{tran2018closer} as visual feature extractor. All these three baselines concatenate the extracted visual and textual features to obtain a combined representation of the video and the question pair, feeding it to a multilayer perceptron network (MLP), followed by a linear layer generating scores for the answers.
 
\noindent\textbf{Memory, Attention, and Composition (MAC)} model \citep{hudson2018} is a compositional visual reasoning model. It decomposes the reasoning task into a series of attention-guided processing steps by isolating memory and control functions from each other. The attention mechanism considers visual and textual features jointly, which leads to robust encodings of the question and the image. Similar to the LSTM-CNN baselines, \textbf{MAC-F} looks at only the first frame, and \mbox{\textbf{MAC-L}} only pays attention to the last frame. \textbf{MAC-V} baseline extends the MAC model by considering the video frames sampled from the given video as the visual input. Like LSTM-CNN-V model, MAC-V also processes videos using R3D. Unlike its non-temporal variations, MAC-F and MAC-L, where the read unit originally has spatial attention over the image, this temporal variation has a read unit that applies spatio-temporal attention over the features extracted from the entire video. 

\noindent \textbf{TVQA} is a multi-stream state-of-the-art video QA neural model \citep{lei2018tvqa}. To adapt this model to our dataset, we only use its video stream branch and omit the answer input by generating scores for the entire answer vocabulary. In parallel with other baselines, TVQA model also extracts visual features by using ResNet-18. Different from the original implementation, our TVQA implementation uses LSTM networks with 256 units, uses a MLP network with 2 layers. Unlike the original model, we do not use GloVe word embeddings \cite{pennington2014glove} to make a fair comparison with the remaining baseline models. 

\noindent \textbf{TVQA+} is another multi-stream video question answering model, which is built upon TVQA model. In contrast to TVQA, TVQA+ uses convolutional networks as sequence encoder instead of LSTM networks, replaces GloVe word embeddings with BERT embeddings \cite{devlin-etal-2019-bert}, and implements a span proposal / prediction mechanism. We do not implement span proposal mechanism, and omit using BERT embeddings to compare TVQA+ with others more fairly as we disable GloVe embeddings in TVQA. Our TVQA+ implementation uses 256 hidden units in all submodules throughout the network, and it generates answer scores by feeding weighted average of fused multi-modal simulation-question representation into a linear layer.

\renewcommand{\arraystretch}{1.15}
\begin{table*}[!t]
\caption{Performances of the tested models on the test set of the CRAFT dataset on easy and hard splits. C, CF, and D columns stand for \emph{Causal}, \emph{Counterfactual}, and \emph{Descriptive} tasks, respectively.}
    \label{table:results}
    \centering
    \resizebox{0.83\textwidth}{!}{
    \begin{tabular}{ll@{$\qquad$} cccc @{$\quad\;$}c@{$\quad\;$} cccc}
    \toprule
    \multirow{2}{*}{} & \multirow{2}{*}{\bf{Model}} & \multicolumn{4}{c}{\bf{Easy Setting}} & &  \multicolumn{4}{c}{\bf{Hard Setting}} \\ \cline{3-6} \cline{8-11}
    & & \bf{C} & \bf{CF} & \bf{D} & \bf{All }& & \bf{C} & \bf{CF} & \bf{D} & \bf{All} \\
    \midrule
    \multirow{4}{*}{Heuristic}& Random & 5.95 & 5.25 & 5.09 & 5.24 & & 5.37 & 4.62 & 5.08 & 4.98 \\
    & AT-Random & 36.67 & 44.34 & 33.95 & 37.47 & & 33.67 & 46.06 & 34.16 & 37.52 \\
     & MFA & 32.68 & 43.28 & 23.53 & 30.72 & & 30.09 & 43.94 & 23.20 & 29.98 \\
    & AT-MFA & 49.62 & 47.21 & 37.57 & 42.03 & & 49.28 & 47.17 & 36.55 & 41.12 \\
    \midrule
    \multirow{2}{*}{Text-only} & LSTM & 53.04 & 53.14 & 38.29 & 44.69 & & 52.51 & 56.24 & 37.25 & 44.52 \\
    & BERT & 48.43 & 50.59 & 37.55 & 42.90 & & 49.28 & 52.12 & 36.52 & 42.52 \\
    
    \midrule
    & LSTM-CNN-F & 53.11 & 55.23 & 44.86 & 49.07 & & 48.07 & 48.12 & 35.54 & 40.64 \\
    Single & LSTM-CNN-L & 54.86 & 55.63 & 43.12 & 48.42 & & 49.86 & 54.44 & 38.88 & 44.66 \\
    Frame & MAC-F & 53.18 & 52.88 & 44.40 & 48.10 & & 51.86 & 53.5 & 42.12 & 46.55 \\
    & MAC-L & 49.97 & 53.08 & 44.54 & 47.83 & & 50.21 & 53.8 & 41.46 & 46.05 \\
    
    \midrule
    & LSTM-CNN-V & 54.65 & 61.42 & 48.12 & 53.01 & & 51.86 & 54.89 & 41.36 & 46.50 \\
    & MAC-V & 53.95 & 57.72 & 44.51 & 49.74 & & 51.22 & 54.71 & 42.94 & 47.31 \\
    Video & TVQA & 53.67 & 55.57 & 36.89 & 44.71 & & 51.00 & 55.12 & 36.31 & 43.46 \\
    & TVQA+ & 54.86 & 60.02 & 40.22 & 48.11 & & 51.00 & 55.12 & 39.09 & 45.12 \\
    & G-SWM & 53.54 & 55.29 & 37.05 & 44.69 & & 51.00 & 48.68 & 37.77 & 42.47 \\
    \midrule
    \multirow{2}{*}{Oracle} & LSTM-D & 51.71 & 55.89 & 63.22 & 59.53 & & 51.93 & 56.00 & 59.57 & 57.64 \\
    & BERT-D & 68.44 & 80.05 & 93.41 & 86.20 & & 66.33 & 79.34 & 91.30 & 84.90 \\
    \midrule
    & & \multicolumn{2}{c}{\bf{C}} & \multicolumn{2}{c}{\bf{CF}} & & \multicolumn{2}{c}{\bf{D}} & \multicolumn{2}{c}{\bf{All}}\\ \cline{3-11}
    & Human & \multicolumn{2}{c}{71.27} & \multicolumn{2}{c}{83.07} & & \multicolumn{2}{c}{87.45} & \multicolumn{2}{c}{76.60} \\
    \bottomrule
    \end{tabular}
   }
\end{table*}

\noindent \textbf{G-SWM} is a recenty proposed object-centric model \citep{lin2020improving}, which is originally designed for simulating possible futures in a scene consisting of multiple dynamic objects. It models each frame in a video by two different latent variables encoding object and context features. We modify G-SWM to solve the reasoning tasks in CRAFT. In particular, our version of G-SWM takes in video frames resized to $64\times 64$ pixels and extracts an object-centric representation of the input video thorough object and context features. These latent codes are then combined and concatenated with the LSTM-based question representation, similar to LSTM-CNN model, just before the final classifier layer.
 
\noindent \textbf{LSTM-D} and \textbf{BERT-D} are \emph{oracle} text-only baselines, which take the natural language description of the causal graph of the simulation (see Figure \ref{casualgraph}) as input in addition to the question. We generate these descriptions from simplified versions of the causal graphs by only considering the \textit{Start}, \textit{End}, \textit{Collision} and \textit{Enter Basket} events, and excluding those involving certain static objects (walls, platforms, ramps, and static balls) which are not mentioned in the questions. 
We first sort the events by their timestamps and concatenate a template-based description of each event to generate the summary. LSTM-D uses two separate LSTM networks process the question and the description, and then a linear layer predicts the answer for the input question/description pair. BERT-D extends the BERT baseline by using the descriptions as prefixes for the input questions.

\subsection{Results}
In Table~\ref{table:results}, we present the performances of the tested models for each question type, considering both the easy and the hard settings explained in Section~\ref{sec:craft-dataset}. As expected, the text only models perform the worst as they completely ignore the visual information present in the videos. Moreover, the performances of the single frame methods are typically lower than those of the video models, showing the importance of the temporal aspect of the questions that a single snapshot of the simulation does not carry enough information. 

As can be seen from Table~\ref{table:results}, there exists a substantial gap between the model performances in the easy and hard settings of CRAFT. Not surprisingly, this is not the case for the text-based baselines, in which it is not important whether a scene layout has been seen before during training or not. Overall, these results suggest that our tested multimodal methods are not able to generalize well to previously unseen scenes. They cannot fully detect the physical interactions and localize the events taking place in a video.

It is worth mentioning that the performances of the models vary between different question types in CRAFT. Out of the three question types, the models consistently perform poorly on the Descriptive questions in that the accuracies are around 23.5\%-48.12\% in the easy setting and 23.2\%-42.9\% in the hard setting. The reason behind this could be attributed to the variety of the answers in this task as it includes questions covering both count, shape, and color of the object(s) (see Figure~\ref{stats}). On the other hand, the accuracies of the models on the remaining questions types are between 32.7\% and 61.4\% in the easy setting, and 30.1\% and 56.2\% in the hard setting.

LSTM-CNN-V baseline does reasonably well on the easy setting, but its generalization capability on the hard setting is not that good. TVQA performs worse than the LSTM-CNN-V baseline, which shows that it is more tailor-fit to video question answering about TV clips, and its performance degrades when it does not have access to subtitles or the related concept detectors. Notably, MAC variants perform the best in the hard setting. MAC model, together with G-SWM, is a more expressive model specifically designed for compositional visual reasoning. G-SWM, however, performs poorly in our experiments, which might be because the scenes in CRAFT usually consist of many objects, thus making it harder to learn decomposing a video into objects and background. This may be resolved by switching to a two-stage framework, in which G-SWM is pretrained first to improve its decomposition ability. For now, we left this as future~work. 

To support our thesis that CRAFT is designed to be easy for humans, but difficult for machines, we also conducted a small human study. We asked 481 randomly selected CRAFT questions to 101 adults. We divided the questions into 5 parts with counterbalancing and every participant took one of the parts randomly. Among these 94 participants, we only considered the ones who responded at least 75\% of the questions, which corresponds to 56 people. As can be seen from Table~\ref{table:results}, there is a large gap ($>29\%$) between human subjects and neural baselines in the hard setting. 

Our oracle models, LSTM-D and BERT-D, perform better than all the tested neural models. Interestingly, the performance of BERT-D is very close to human performance, even slightly outperforming humans for the descriptive questions. Clearly, to excel in this task, a model must capture the interactions between the dynamic objects with each other and with the environment. 

\section{Conclusion}
We have presented CRAFT, a new VQA dataset to test causal reasoning capabilities of the current models. Motivated by the Force Dynamics Theory, which highlights distinct causative verbs, CRAFT requires models to perform temporal and causal reasoning and even to imagine alternative versions  of the events occurring in videos. Our results demonstrate that, while human adults can reason about the physical interactions between objects, these questions cannot be solved reliably by current models. At present, there is substantial room for improvement compared to humans. In our experiments, we did not report the results of recent neuro-symbolic models, e.g. NS-DR~\citep{yi2019clevrer}. Such approaches are very compelling and worth pursuing, but they currently require extra object-level annotations. Another exciting direction is to test object-centric methods other than G-SWM. However, it seems that they might require extra pretraining or self-supervised objectives, as explored by~\citet{ding2020}. We believe that developing more effective models for CRAFT is an exciting research direction for video QA systems to mimic humans in causal reasoning about forces and interactions. 

\section*{Acknowledgments}
This work was supported in part by GEBIP 2018 Award of the Turkish Academy of Sciences to E. Erdem and T. Goksun, BAGEP 2021 Award of the Science Academy to A. Erdem, and AI Fellowship to Ilker Kesen provided by the KUIS AI Center.

\bibliographystyle{acl_natbib} 
\bibliography{craft} 


\clearpage
\onecolumn
\appendix
\section{Appendix}
\counterwithin{figure}{section}
\counterwithin{table}{section}
\setcounter{figure}{0}  
\setcounter{table}{0}  

\subsection{Functional Modules}
\label{app_funcmodules}

CRAFT questions are represented with functional programs. Input and output types for our functional modules are listed in Table~\ref{table:functionalModulesInputOutputTypes}. Lists of all functional modules are also provided in Tables \ref{table:inputFunctionalModules}-\ref{table:auxFunctionalModules}.

\begin{table}[!ht]
  \caption{Input and output types of functional modules in CRAFT.}
  \label{table:functionalModulesInputOutputTypes}
  \centering
  \begin{tabular}{ll}
    \toprule
    \cmidrule(r){1-2}
    \textbf{Type} & \textbf{Description} \\
    \midrule
    \textit{Object} & A dictionary holding static and dynamic attributes of an object \\
    \textit{ObjectSet} & A list of unique objects \\
    \textit{ObjectSetList} & A list of \textit{ObjectSet} \\
    \textit{Event} & A dictionary holding information of a specific event \\
    \textit{EventSet} & A list of unique events \\
    \textit{EventSetList} & A list of \textit{EventSet} \\
    \textit{Size} & A tag indicating the size of an object \\
    \textit{Color} & A tag indicating the color of an object \\
    \textit{Shape} & A tag indicating the shape of an object \\
    \textit{Integer} & Standard integer type \\
    \textit{Bool} & Standard boolean type \\
    \textit{BoolList} & A list of \textit{Bool} \\
    \bottomrule
  \end{tabular}
\end{table}

\renewcommand{\arraystretch}{1.5} 
\begin{table}[!ht]
  \caption{Input functional modules in CRAFT.}
  \label{table:inputFunctionalModules}
  \centering
 \resizebox{0.9\linewidth}{!}{
  \begin{tabular}{lp{5.5cm}ll}
    \toprule
    \cmidrule(r){1-2}
    \textbf{Module} & \textbf{Description} & \textbf{Input Types} & \textbf{Output Type} \\
    \midrule
    \texttt{SceneAtStart} & Returns the attributes of all objects at the start of the simulation & \textit{None} & \textit{ObjectSet} \\
    \texttt{SceneAtEnd} & Returns the atttributes of all objects at the end of the simulation & \textit{None} & \textit{ObjectSet} \\
    \texttt{StartSceneStep} & Returns \texttt{0} & \textit{None} & \textit{Integer} \\
    \texttt{EndSceneStep} & Returns \texttt{-1} & \textit{None} & \textit{Integer} \\
    \texttt{Events} & Returns all of the events happening between the start and the end of the simulation & \textit{None} & \textit{EventSet} \\
    \bottomrule
  \end{tabular}
  }
\end{table}

\renewcommand{\arraystretch}{1.5} 
\begin{table}[!ht]
  \caption{Output functional modules in CRAFT.}
  \label{table:outputFunctionalModules}
  \centering
  \resizebox{0.9\linewidth}{!}{
  \begin{tabular}{lp{5.5cm}ll}
    \toprule
    \cmidrule(r){1-2}
    \textbf{Module} & \textbf{Description} & \textbf{Input Types} & \textbf{Output Type} \\
    \midrule
    \texttt{QueryColor} & Returns the color of the input object  & \textit{Object} & \textit{Color} \\
    \texttt{QueryShape} & Returns the shape of the input object  & \textit{Object} & \textit{Shape} \\
    \texttt{Count} & Returns the size of the input list  & \textit{ObjectSet} & \textit{Integer} \\
    \texttt{Exist} & Returns true if the input list is not empty  & \textit{ObjectSet} / \textit{EventSet} & \textit{Bool} \\
    \texttt{AnyFalse} & Returns true if there is at least one false in a boolean list & \textit{BoolList} & \textit{Bool} \\
    \texttt{AnyTrue} & Returns true if there is at least one true in a boolean list & \textit{BoolList} & \textit{Bool} \\
    \texttt{IsBefore} & Returns whether the first event happened before the second event & \textit{(Event, Event)} & \textit{Bool} \\
    \texttt{IsAfter} & Returns whether the first event happened after the second event & \textit{(Event, Event)} & \textit{Bool} \\
    \bottomrule \\ & & &\\
  \end{tabular}
  }
\end{table}

\renewcommand{\arraystretch}{1.5} 
\begin{table}[!ht]
  \caption{Object filter functional modules in CRAFT.}
  \label{table:objectFilteringFunctionalModules}
  \centering
  \resizebox{0.9\linewidth}{!}{
  \begin{tabular}{lp{5.5cm}ll}
    \toprule
    \cmidrule(r){1-2}
    \textbf{Module} & \textbf{Description} & \textbf{Input Types} & \textbf{Output Type} \\
    \midrule
    \texttt{FilterColor} & Returns the list of objects which have a color same with the input color & \textit{(ObjectSet, Color)} & \textit{ObjectSet} \\
    \texttt{FilterShape} & Returns the list ofobjects which have a shape same with the input shape & \textit{(ObjectSet, Shape)} & \textit{ObjectSet} \\
    \texttt{FilterSize} & Returns the list of objects which have a size same with the input size & \textit{(ObjectSet, Size)} & \textit{ObjectSet} \\
    \texttt{FilterDynamic} & Returns the list of dynamic objects from an object set & \textit{ObjectSet} & \textit{ObjectSet} \\
    \texttt{FilterMoving} & Returns the list of objects that are in motion at the step specified & \textit{(ObjectSet, Integer)} & \textit{ObjectSet} \\
    \texttt{FilterStationary} & Returns the list of objects that are stationary at the step specified & \textit{(ObjectSet, Integer)} & \textit{ObjectSet} \\
    \bottomrule
  \end{tabular}
  }
\end{table}

\renewcommand{\arraystretch}{1.5} 
\begin{table}[!ht]
  \caption{Event filter functional modules in CRAFT.}
  \label{table:eventFilteringFunctionalModules}
  \centering
  \resizebox{\linewidth}{!}{%
  \begin{tabular}{lp{5.5cm}ll}
    \toprule
    \cmidrule(r){1-2}
    \textbf{Module} & \textbf{Description} & \textbf{Input Types} & \textbf{Output Type} \\
    \midrule
    \texttt{FilterEvents} & Returns the list of events about a specific object from an event set & \textit{(EventSet, Object)} & \textit{EventSet} \\
    \texttt{FilterCollision} & Returns the list of collision events from an event set& \textit{EventSet} & \textit{EventSet} \\
    \texttt{FilterCollisionWithDynamics} & Returns the list of collision events involving dynamic objects & \textit{EventSet} & \textit{EventSet} \\
    \texttt{FilterCollideGround} & Returns the list of collision events involving the ground & \textit{EventSet} & \textit{EventSet} \\
    \texttt{FilterCollideGroundList} & Returns the list of collision event sets involving the ground & \textit{EventSetList} & \textit{EventSetList} \\
    \texttt{FilterCollideBasket} & Returns the list of collision events involving the basket & \textit{EventSet} & \textit{EventSet} \\
    \texttt{FilterCollideBasketList} & Returns the list of collision event sets involving the basket & \textit{EventSetList} & \textit{EventSetList} \\
    \texttt{FilterEnterBasket} & Returns the In Basket events & \textit{EventSet} & \textit{EventSet} \\
    \texttt{FilterEnterBasketList} & Returns the list of In Basket event sets & \textit{EventSetList} & \textit{EventSetList} \\
    \texttt{FilterBefore} & Returns the events from the input list that happens before input event & \textit{(EventSet, Event)} & \textit{EventSet} \\
    \texttt{FilterAfter} & Returns the events from the input list that happened after input event & \textit{(EventSet, Event)} & \textit{EventSet} \\
    \texttt{FilterFirst} & Returns the first event & \textit{EventSet} & \textit{Event} \\
    \texttt{FilterLast} & Returns the last event & \textit{EventSet} & \textit{Event} \\
    \texttt{EventPartner} & Returns the object interacting with the input object through the specified event & \textit{(Event, Object)} & \textit{Object} \\
    \texttt{FilterObjectsFromEvents} & Returns the objects from the specified events & \textit{EventSet} & \textit{ObjectSet} \\
    \texttt{FilterObjectsFromEventsList} & Returns the list of object sets from a list of event sets & \textit{EventSetList} & \textit{ObjectSetList} \\
    \texttt{GetCounterfactEvents} & Returns the event list if a specific object is removed from the scene & \textit{Object} & \textit{EventSet} \\
    \texttt{GetCounterfactEventsList} & Returns the counterfactual event list for all objects in an object set & \textit{ObjectSet} & \textit{EventSetList} \\
    \bottomrule
  \end{tabular}
  }
\end{table}

\clearpage

\renewcommand{\arraystretch}{1.5} 
\begin{table}[!ht]
  \caption{Auxiliary functional modules in CRAFT.}
  \label{table:auxFunctionalModules}
  \centering
  \resizebox{0.9\linewidth}{!}{
  \begin{tabular}{lp{5.5cm}ll}
    \toprule
    \cmidrule(r){1-2}
    \textbf{Module} & \textbf{Description} & \textbf{Input Types} & \textbf{Output Type} \\
    \midrule
    \texttt{Unique} & Returns the single object from the input list, if the list has multiple elements returns \texttt{INVALID} & \textit{ObjectSet} & \textit{Object} \\
    \texttt{Intersect} & Applies the set intersection operation & \textit{(ObjectSet, ObjectSet)} & \textit{ObjectSet} \\
    \texttt{IntersectList} & Intersects an object set with multiple object sets & \textit{(ObjectSetList, ObjectSet)} & \textit{ObjectSetList} \\
    \texttt{Difference} & Applies the set difference operation & \textit{(ObjectSet, ObjectSet)} & \textit{ObjectSet} \\
    \texttt{ExistList} & Applies the Exist operation to each item in the input list returning a boolean list & \textit{ObjectSetList} / \textit{EventSetList} & \textit{BoolList} \\
    \texttt{AsList} & Returns an object set containing a single element specified by the input object & \textit{Object} & \textit{ObjectSet} \\
    \bottomrule \\ & & & \\

  \end{tabular}
  }
\end{table}

\subsection{Example Programs}
\label{app_programs}
Here we provide example functional programs for some of the sample questions provided in Figure~\ref{svqa_eg1}, which are used to extract the correct answers using our simulation environment. Figures~\ref{desc_programs}~to~\ref{prevent_programs} provide functional program samples that are designed for CRAFT descriptive, counterfactual, cause, enable, and prevent questions, respectively.

\begin{figure}[!ht]
\textbf{Question}: \textit{"How many objects fall to the ground?"}\\
\\
{
\renewcommand{\arraystretch}{1.0}
\begin{tabular}{l}
\texttt{Count (}\\
\hspace*{1cm} \texttt{FilterDynamic (}\\
\hspace*{2cm} \texttt{FilterObjectsFromEvents (}\\
\hspace*{3cm} \texttt{FilterCollideGround (}\\
\hspace*{4cm} \texttt{Events ()}\\
\hspace*{3cm} \texttt{)}\\
\hspace*{2cm} \texttt{)}\\
\hspace*{1cm} \texttt{)}\\
\texttt{)}\\
\texttt{}\\
\end{tabular}
}
\\
\textbf{Question}: \textit{"After entering the basket, does the small yellow square collide with other objects?"}\\
\\
{
\scriptsize
\renewcommand{\arraystretch}{1.0}
\resizebox{\linewidth}{!}{
\begin{tabular}{l}
\texttt{Var QueryObject = FilterShape ( FilterColor ( FilterSize ( SceneAtStart(),  "Small" ) , "Yellow"), "Cube" )
}\\
\texttt{Var SmallYellowCubeEvents = FilterEvents ( Events(), QueryObject ) }\\

\texttt{Exist (}\\
\hspace*{1cm} \texttt{FilterAfter (}\\
\hspace*{2cm} \texttt{FilterCollisionWithDynamics ( SmallYellowCubeEvents ),}\\
\hspace*{3cm} \texttt{FilterFirst (}\\
\hspace*{4cm} \texttt{FilterEnterBasket ( SmallYellowCubeEvents ) }\\
\hspace*{3cm} \texttt{)}\\
\hspace*{2cm} \texttt{)}\\
\hspace*{1cm} \texttt{)}\\
\texttt{)}\\
\end{tabular}}
}
\caption{Example programs for \emph{descriptive} questions.}
\label{desc_programs}
\end{figure}

\begin{figure}[!t]
\textbf{Question}: \textit{"How many objects fall to the ground if the small yellow box is removed?"}\\
\\
{
\scriptsize
\renewcommand{\arraystretch}{1.0}
\resizebox{\linewidth}{!}{
\begin{tabular}{l}
\texttt{Var QueryObject = FilterShape ( FilterColor ( FilterSize ( SceneAtStart(),  "Small" ) , "Yellow"), "Cube" )}\\
\texttt{Count (}\\
\hspace*{1cm} \texttt{FilterObjectsFromEvents (}\\
\hspace*{2cm} \texttt{FilterCollideGround (}\\
\hspace*{3cm} \texttt{GetCounterfactEvents ( QueryObject ) }\\
\hspace*{2cm} \texttt{)}\\
\hspace*{1cm} \texttt{)}\\
\texttt{)}\\
\texttt{}\\
\end{tabular}
}}
\\
\\
\textbf{Question}: \textit{"Will the small gray box enter the basket if any of the other objects are removed?"}\\
\\
{
\scriptsize
\renewcommand{\arraystretch}{1.0}
\begin{tabular}{l}
\texttt{Var QueryObject = FilterShape ( FilterColor ( FilterSize ( SceneAtStart(),  "Small" ) , "Gray"), "Cube" )}\\
\texttt{Var OtherDynamicObjects = Difference ( FilterDynamic ( SceneAtStart() ), AsList ( QueryObject ) )}\\
\texttt{AnyTrue (}\\
\hspace*{1cm} \texttt{ExistList (}\\
\hspace*{2cm} \texttt{IntersectList (}\\
\hspace*{3cm} \texttt{FilterObjectsFromEventsList (}\\
\hspace*{4cm} \texttt{FilterEnterBasketList (}\\
\hspace*{5cm} \texttt{GetCounterfactEventsList ( OtherDynamicObjects )}\\
\hspace*{4cm} \texttt{)}\\
\hspace*{3cm} \texttt{),}\\
\hspace*{3cm} \texttt{AsList (}\\
\hspace*{4cm} \texttt{QueryObject}\\
\hspace*{3cm} \texttt{)}\\
\hspace*{2cm} \texttt{)}\\
\hspace*{1cm} \texttt{)}\\
\texttt{)}\\
\end{tabular}
}
\caption{Example programs for \emph{counterfactual} questions.}
\label{count_programs}
\end{figure}

\begin{figure}[!ht]
\textbf{Question}: \textit{"Does the small brown sphere cause the tiny yellow box to enter the basket?"}\\
\\
{
\scriptsize
\renewcommand{\arraystretch}{1.0}
\begin{tabular}{l}
\texttt{Var AffectorObject = FilterShape ( FilterColor ( FilterSize ( SceneAtStart(),  "Small" ) , "Brown"), “Circle” )}\\
\texttt{Var PatientObject = FilterShape ( FilterColor ( FilterSize ( SceneAtStart(),  "Small" ) , "Yellow"), "Cube" )}\\
\texttt{Exist (}\\
\hspace*{1cm} \texttt{FilterStationary (}\\
\hspace*{2cm} \texttt{Intersect (}\\
\hspace*{3cm} \texttt{Difference (}\\
\hspace*{4cm} \texttt{FilterObjectsFromEvents (}\\
\hspace*{5cm} \texttt{FilterEnterBasket (}\\
\hspace*{6cm} \texttt{Events()}\\
\hspace*{5cm} \texttt{)}\\
\hspace*{4cm} \texttt{),}\\
\hspace*{4cm} \texttt{FilterObjectsFromEvents (}\\
\hspace*{5cm} \texttt{FilterEnterBasket (}\\
\hspace*{6cm} \texttt{GetCounterfactEvents (}\\
\hspace*{7cm} \texttt{AffectorObject}\\
\hspace*{6cm} \texttt{)}\\
\hspace*{5cm} \texttt{)}\\
\hspace*{4cm} \texttt{)}\\
\hspace*{3cm} \texttt{),}\\
\hspace*{3cm} \texttt{AsList ( PatientObject )}\\
\hspace*{2cm} \texttt{),}\\
\hspace*{2cm} \texttt{StartSceneStep()}\\
\hspace*{1cm} \texttt{)}\\
\texttt{)}\\
\end{tabular}
}
\caption{Example program for \emph{cause} questions.}
\label{cause_programs}
\end{figure}

\begin{figure}[!ht]
\textbf{Question}: \textit{"How many objects does the small gray block enable to enter the basket?"}\\
\\
{
\scriptsize
\renewcommand{\arraystretch}{1.0}
\resizebox{\linewidth}{!}{\begin{tabular}{l}
\texttt{Var AffectorObject = FilterShape ( FilterColor ( FilterSize ( SceneAtStart(),  "Small" ) , "Gray"), "Cube" )}\\
\texttt{Count (}\\
\hspace*{1cm} \texttt{FilterMoving (}\\
\hspace*{2cm} \texttt{Difference (}\\
\hspace*{3cm} \texttt{Difference (}\\
\hspace*{4cm} \texttt{FilterObjectsFromEvents (}\\
\hspace*{5cm} \texttt{FilterEnterBasket (}\\
\hspace*{6cm} \texttt{Events()}\\
\hspace*{5cm} \texttt{)}\\
\hspace*{4cm} \texttt{),}\\
\hspace*{4cm} \texttt{FilterObjectsFromEvents (}\\
\hspace*{5cm} \texttt{FilterEnterBasket (}\\
\hspace*{6cm} \texttt{GetCounterfactEvents (}\\
\hspace*{7cm} \texttt{AffectorObject}\\
\hspace*{6cm} \texttt{)}\\
\hspace*{5cm} \texttt{)}\\
\hspace*{4cm} \texttt{)}\\
\hspace*{3cm} \texttt{),}\\
\hspace*{3cm} \texttt{AsList ( AffectorObject )}\\
\hspace*{2cm} \texttt{),}\\
\hspace*{2cm} \texttt{StartSceneStep()}\\
\hspace*{1cm} \texttt{)}\\
\texttt{)}\\
\end{tabular}
}}
\caption{Example program for \emph{enable} questions.}
\label{enable_programs}
\end{figure}

\begin{figure}[!ht]
\textbf{Question}: \textit{"Does the small yellow square prevent the tiny brown circle from entering the basket?"}\\
\\
{
\scriptsize
\renewcommand{\arraystretch}{1.0}
\resizebox{\linewidth}{!}{\begin{tabular}{l}
\texttt{Var AffectorObject = FilterShape ( FilterColor ( FilterSize ( SceneAtStart(),  "Small" ) , "Yellow"), "Cube" )}\\
\texttt{Var PatientObject = FilterShape ( FilterColor ( FilterSize ( SceneAtStart(),  "Small" ) , "Brown"), "Circle" )}\\
\texttt{Exist (}\\
\hspace*{1cm} \texttt{FilterMoving (}\\
\hspace*{2cm} \texttt{Intersect (}\\
\hspace*{3cm} \texttt{Difference (}\\
\hspace*{4cm} \texttt{FilterObjectsFromEvents (}\\
\hspace*{5cm} \texttt{FilterEnterBasket (}\\
\hspace*{6cm} \texttt{GetCounterfactEvents (}\\
\hspace*{7cm} \texttt{AffectorObject}\\
\hspace*{6cm} \texttt{)}\\
\hspace*{5cm} \texttt{)}\\
\hspace*{4cm} \texttt{),}\\
\hspace*{4cm} \texttt{FilterObjectsFromEvents (}\\
\hspace*{5cm} \texttt{FilterEnterBasket (}\\
\hspace*{6cm} \texttt{Events()}\\
\hspace*{5cm} \texttt{)}\\
\hspace*{4cm} \texttt{)}\\
\hspace*{3cm} \texttt{),}\\
\hspace*{3cm} \texttt{AsList ( PatientObject )}\\
\hspace*{2cm} \texttt{),}\\
\hspace*{2cm} \texttt{StartSceneStep()}\\
\hspace*{1cm} \texttt{)}\\
\texttt{)}\\
\end{tabular}
}}
\caption{Example program for \emph{prevent} questions.}
\label{prevent_programs}
\end{figure}

\clearpage
\subsection{Implementation Details}

Unless otherwise specified, all learnable baselines are trained with Adam optimizer \citep{kingma2014adam} with default hyperparameters. LSTM and single-frame models are trained for 75 epochs with a batch size of 64. All temporal baselines are trained for 30 epochs with a batch size of 32. G-SWM is trained for 100 epochs using a batch size of 64 with Adam optimizer and a learning rate of 0.0001. Input videos are downsampled at 5 frames per second (fps), and their frames are resized to $112\times 112$ pixels. We used mixed precision strategy to train baselines more efficiently on Tesla V100 and Tesla P4 GPUs, except TVQA+, which is trained using full precision. Training single-frame models takes 2 minutes and training video models approximately 20-30 minutes per epoch. All word embeddings have a length of 256 and are randomly initialized. Pretrained convolutional video and image encoders are jointly trained with the rest of the networks. We use negative log-likelihood loss function for all models where the models predict a distribution over the set of possible answers. All models are tuned based on their performances on the validation split.

\subsection{Detailed Quantitative Results}
In this subsection, we share the quantitative results in more detail for different scenes and question types. Table~\ref{tab:subcategories} describes the subcategories of the question types exist in CRAFT, together with a sample question. Table \ref{table:per-scene-easy-split} and Table \ref{table:per-scene-hard-split} present the results per scene on the easy and hard splits, respectively, and Table \ref{table:question-type-easy-split} and \ref{table:question-type-hard-split} respectively demonstrate the results per question type on the  easy split and hard splits.

\begin{table*}[!ht]
\caption{The question subcategories in the CRAFT dataset.}
    \centering
    \resizebox{\textwidth}{!}{
    \begin{tabular}{lp{7cm}p{7cm}}
    \toprule
    \bf{Subcategory} & \bf{Description} & \bf{Sample Question} \\
    \midrule
    C/A & Yes/no questions that require causal reasoning & \textit{Does the Z C S cause the Z2 C2 S2 to enter the basket?}\\
    C/N & Causal reasoning questions with counting & \textit{What is the number of objects that the Z C S enables to enter the basket?}\\
    CF/N & Counterfactual reasoning with counting & \textit{How many objects enter the basket if the Z C S is removed?}\\
    CF/O & Counterfactual yes/no questions & \textit{Will the Z2 C2 S2 enter the basket if the Z C S is removed?}\\
    D/2Q & Descriptive counting questions about the last state & \textit{How many objects are moving when the video ends?} \\
    D/C & Descriptive questions about the object color & \textit{What color is the object the Z C S last collides with?}\\
    D/C-T & Temporal yes/no questions with respect to a certain event &  \textit{Before falling to the ground, does the Z C S collide with other objects?}\\
    D/N-T & Counting with respect to some reference event & \textit{Before falling to the ground, does the Z C S collide with other objects?}\\
    D/N-V & Descriptive counting questions about events & \textit{How many objects fall to the ground?}\\
    D/S & Descriptive questions about the object shape & \textit{What is the shape of the object the Z C S first collides with?}\\
    D/TO & Temporal yes/no questions about events with respect to an object & \textit{Does the Z C S enter the basket before the Z2 C2 S2 does?}\\
    \bottomrule
    \end{tabular}}
    \label{tab:subcategories}
\end{table*}

\renewcommand{\arraystretch}{1.3}
\begin{table*}[!t]
    \caption{Performances of the tested models per scene on the test set of the easy split of CRAFT.}
    \label{table:per-scene-easy-split}
    \centering
    \resizebox{\textwidth}{!}{
    \begin{tabular}{c@{$\;\;$}l @{$\;\;$}c@{$\;\;$}c@{$\;\;$}c@{$\;\;$}c@{$\;\;$}c@{$\;\;$}c@{$\;\;$}c@{$\;\;$}c@{$\;\;$}c@{$\;\;$}c@{$\;\;$}c@{$\;\;$}c@{$\;\;$}c@{$\;\;$}c@{$\;\;$}c@{$\;\;$}c@{$\;\;$}c@{$\;\;$}c@{$\;\;$}c@{$\;\;$}c}
    \toprule
     & \backslashbox{\bf{Model}}{\bf{Scene}} & 
     \includegraphics[width=0.05\linewidth]{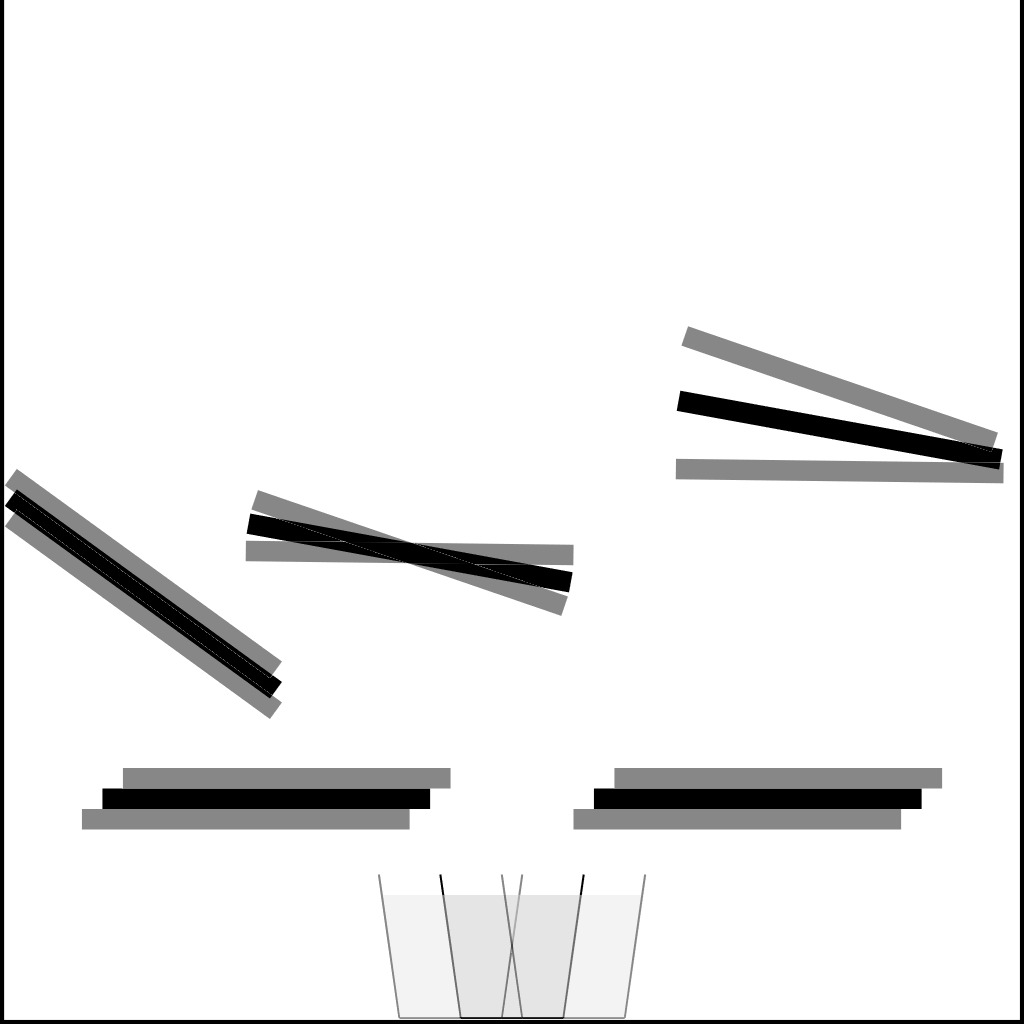} &
     \includegraphics[width=0.05\linewidth]{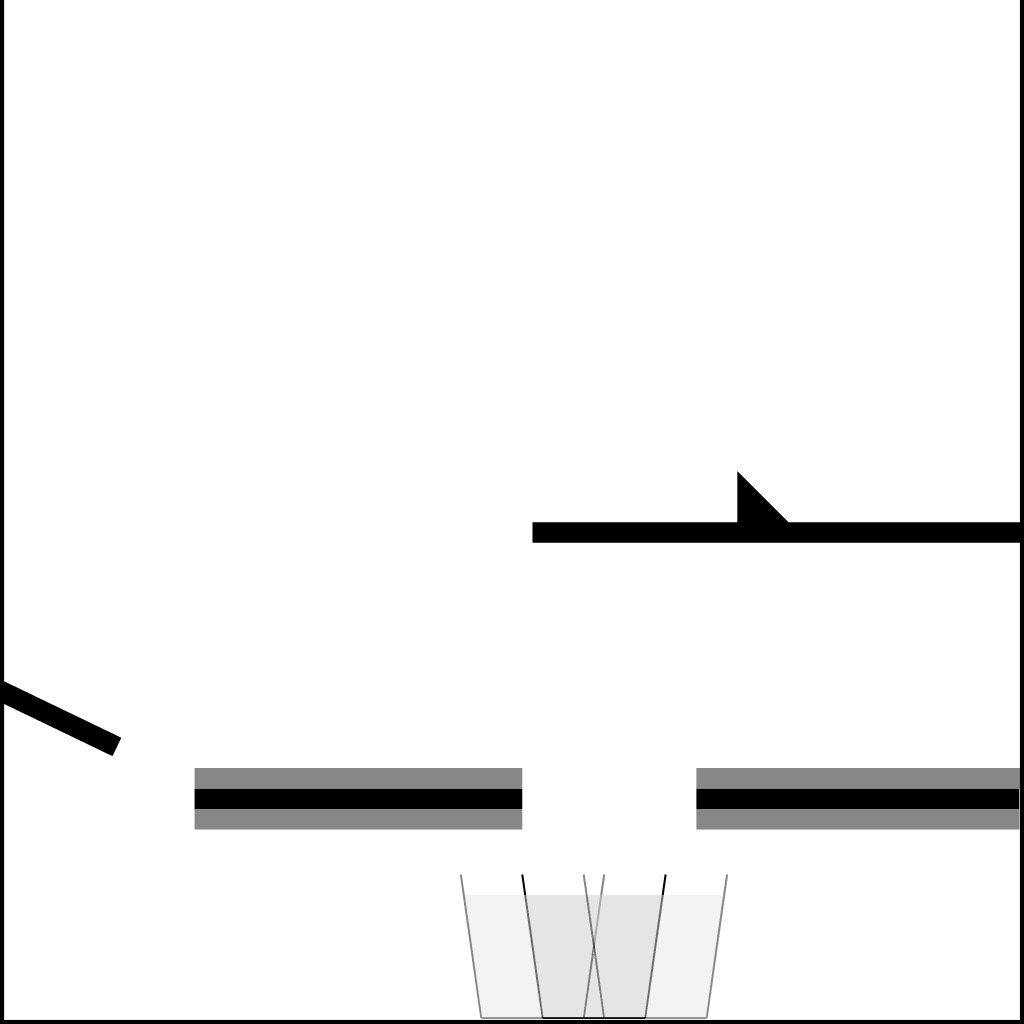} &
     \includegraphics[width=0.05\linewidth]{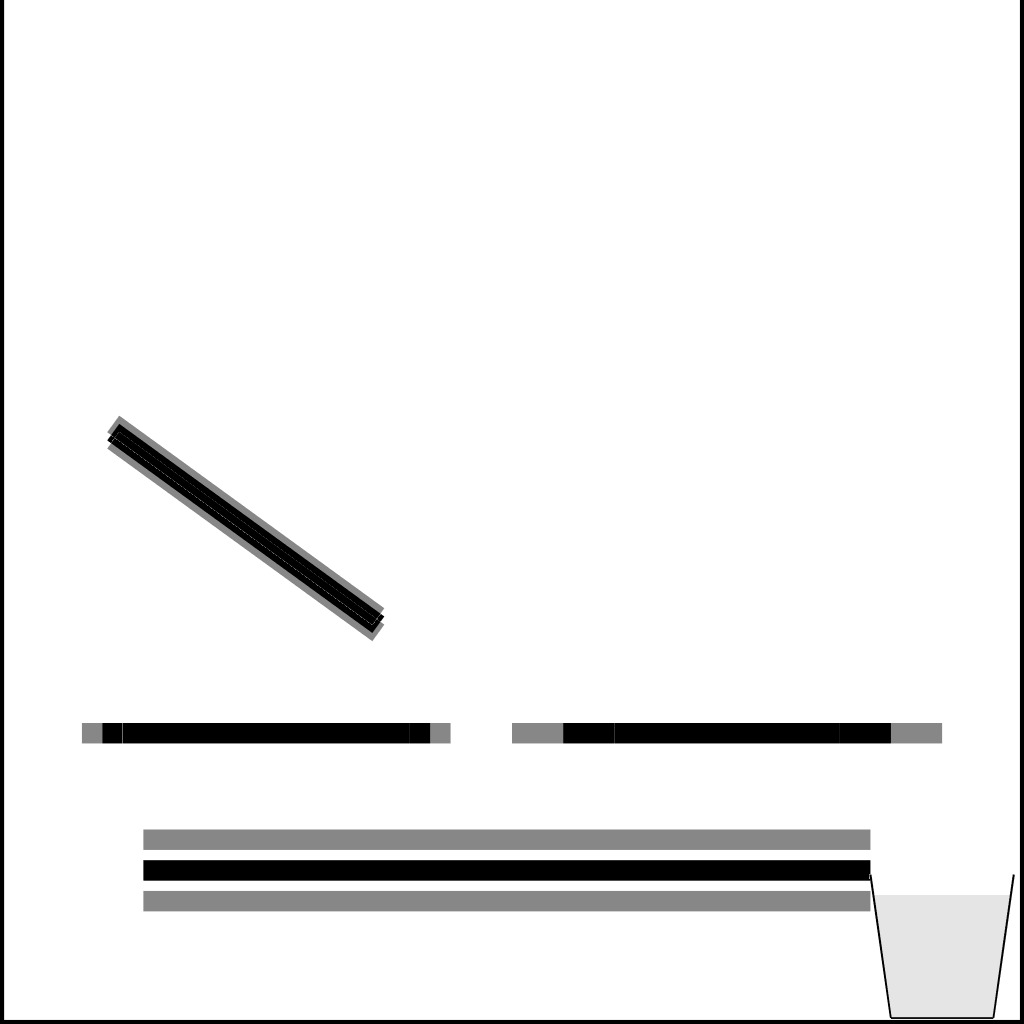} &
     \includegraphics[width=0.05\linewidth]{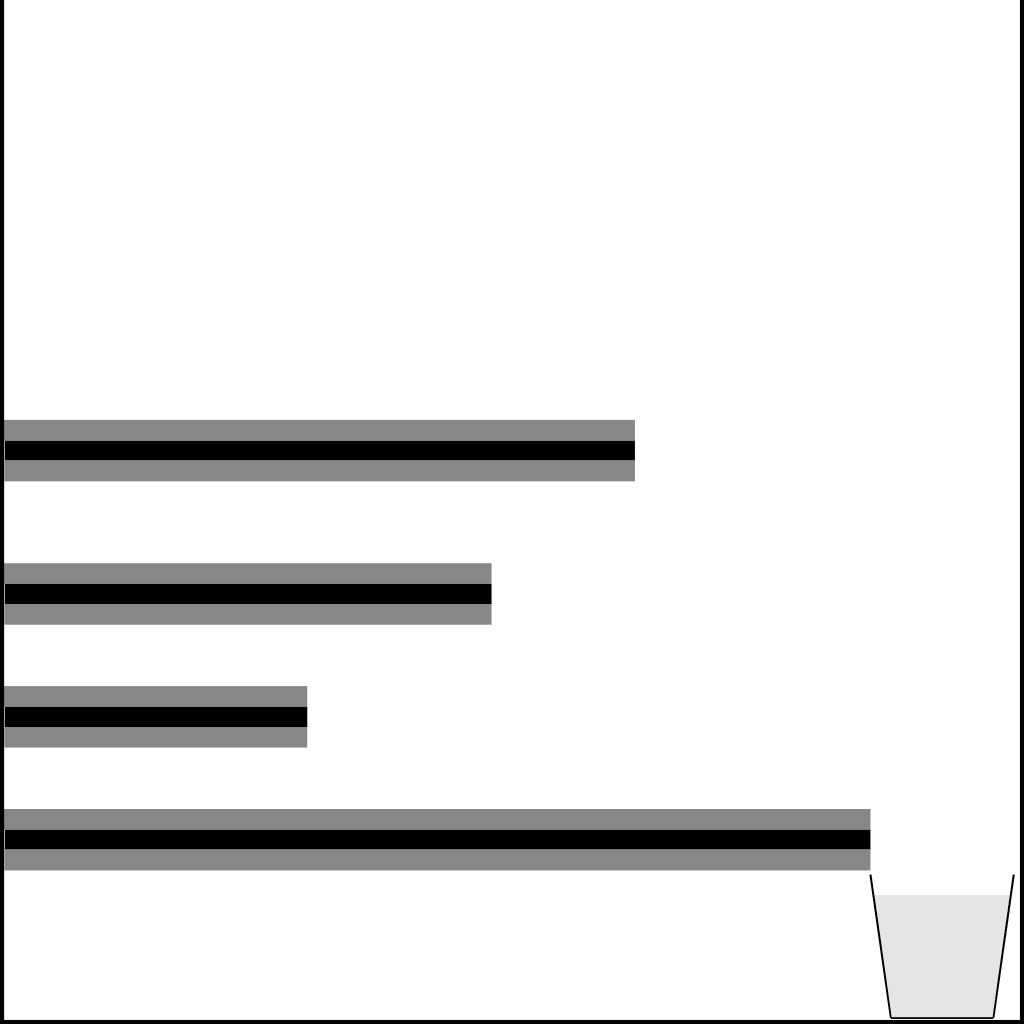} &
     \includegraphics[width=0.05\linewidth]{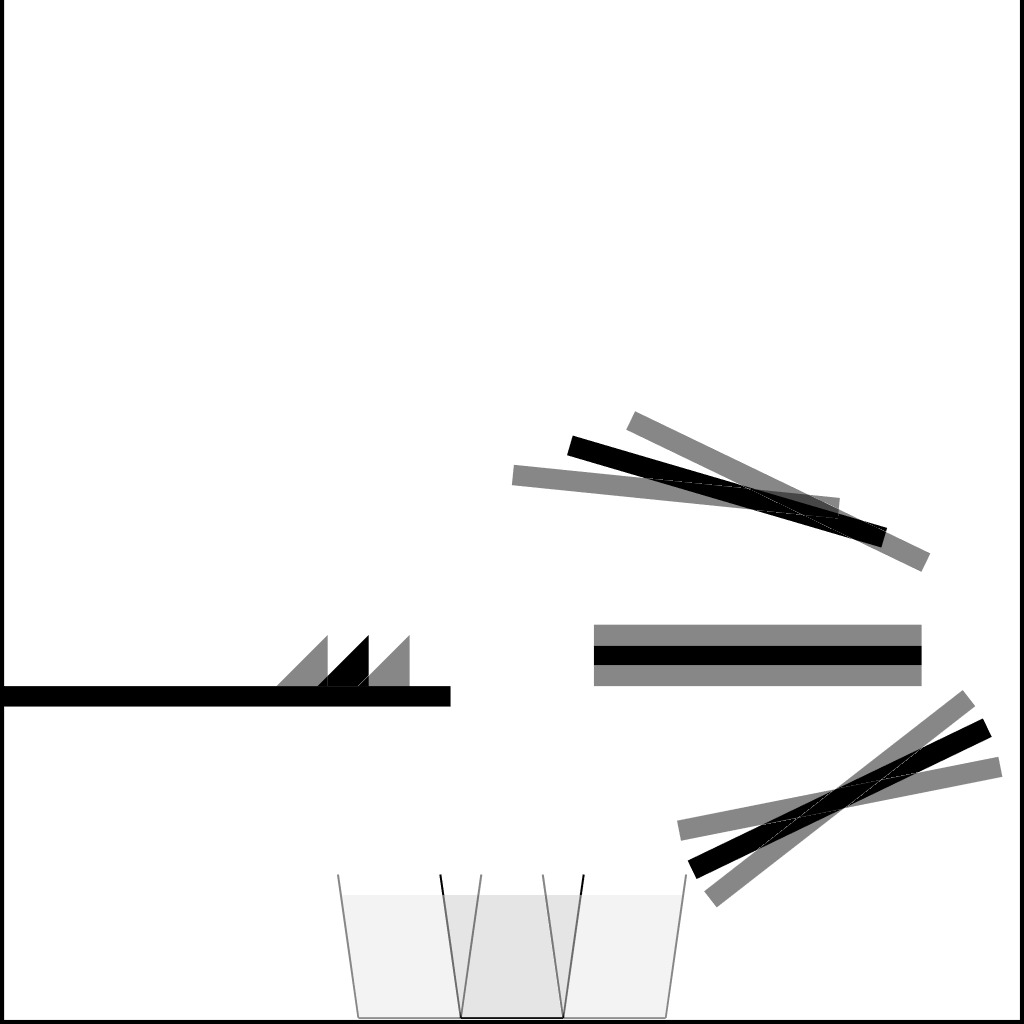} &
     \includegraphics[width=0.05\linewidth]{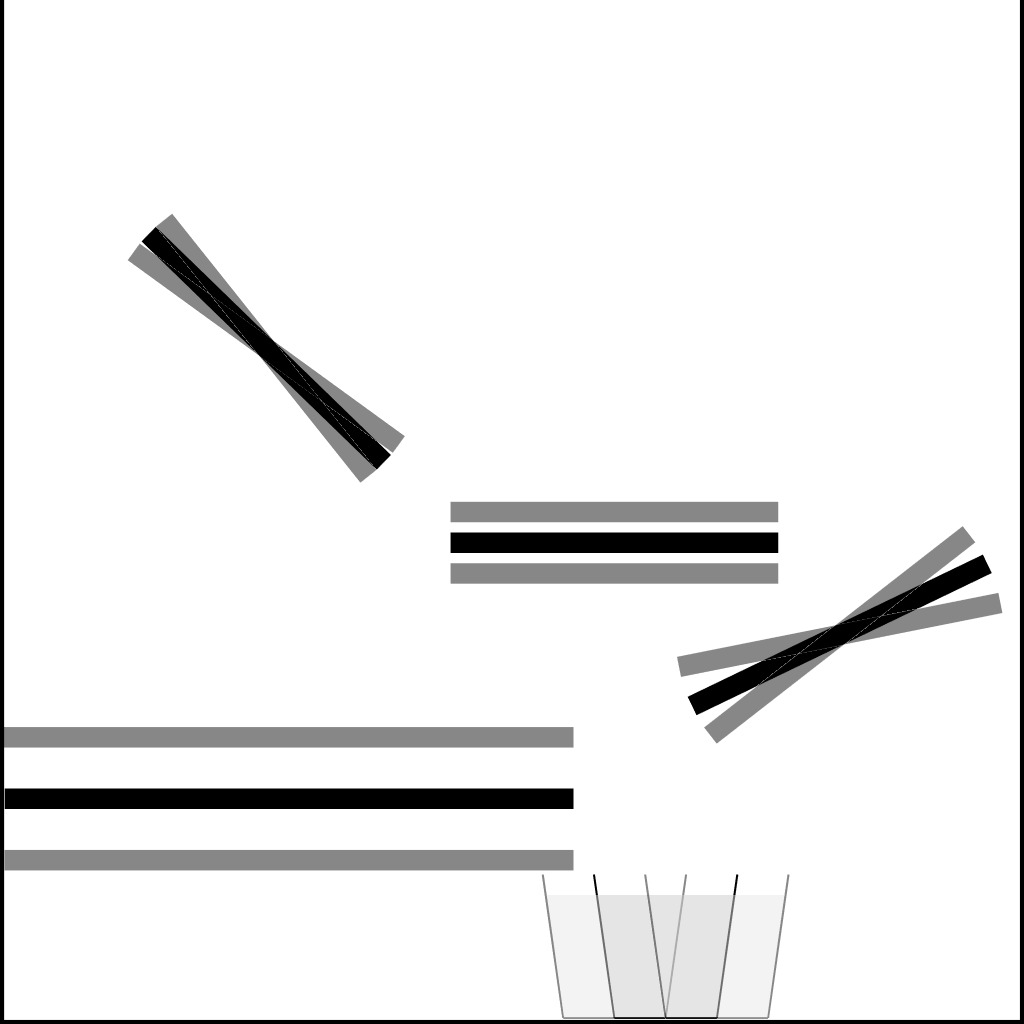} &
     \includegraphics[width=0.05\linewidth]{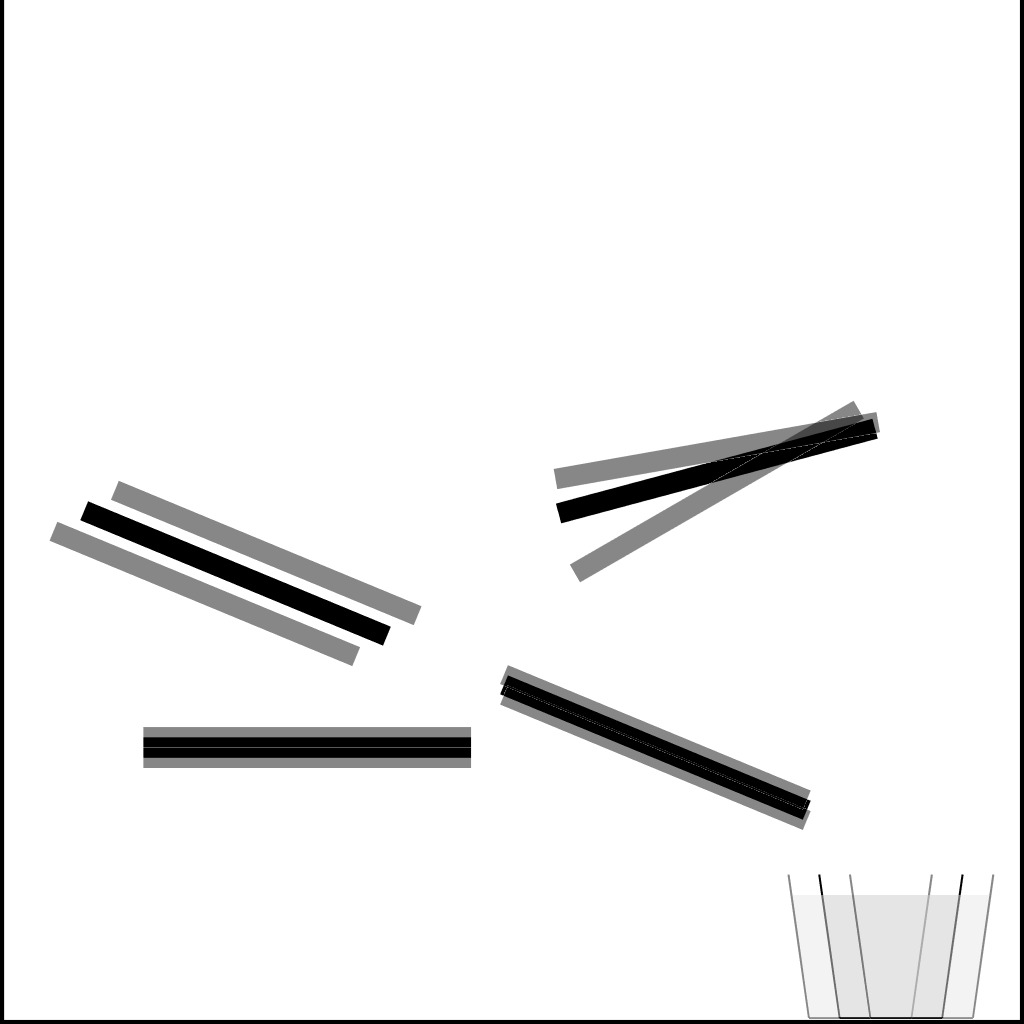} &
     \includegraphics[width=0.05\linewidth]{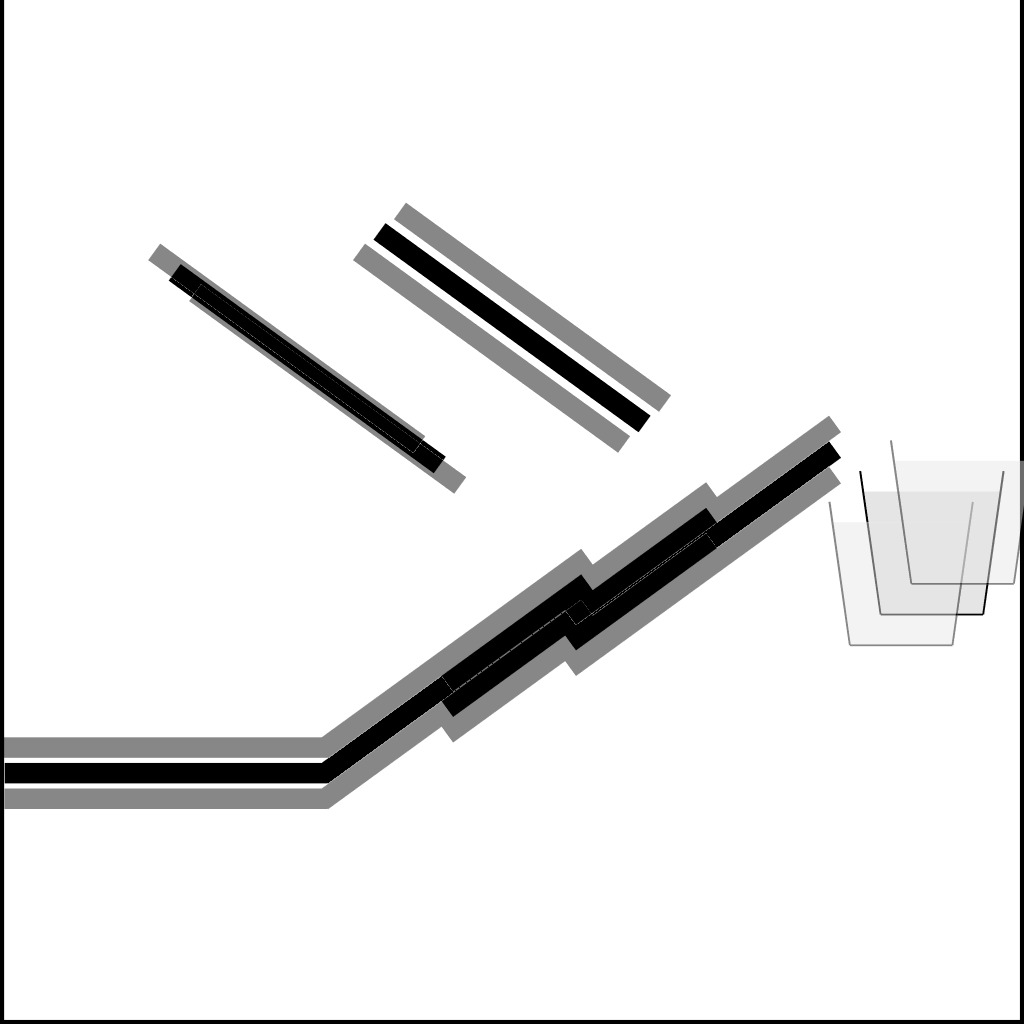} &
     \includegraphics[width=0.05\linewidth]{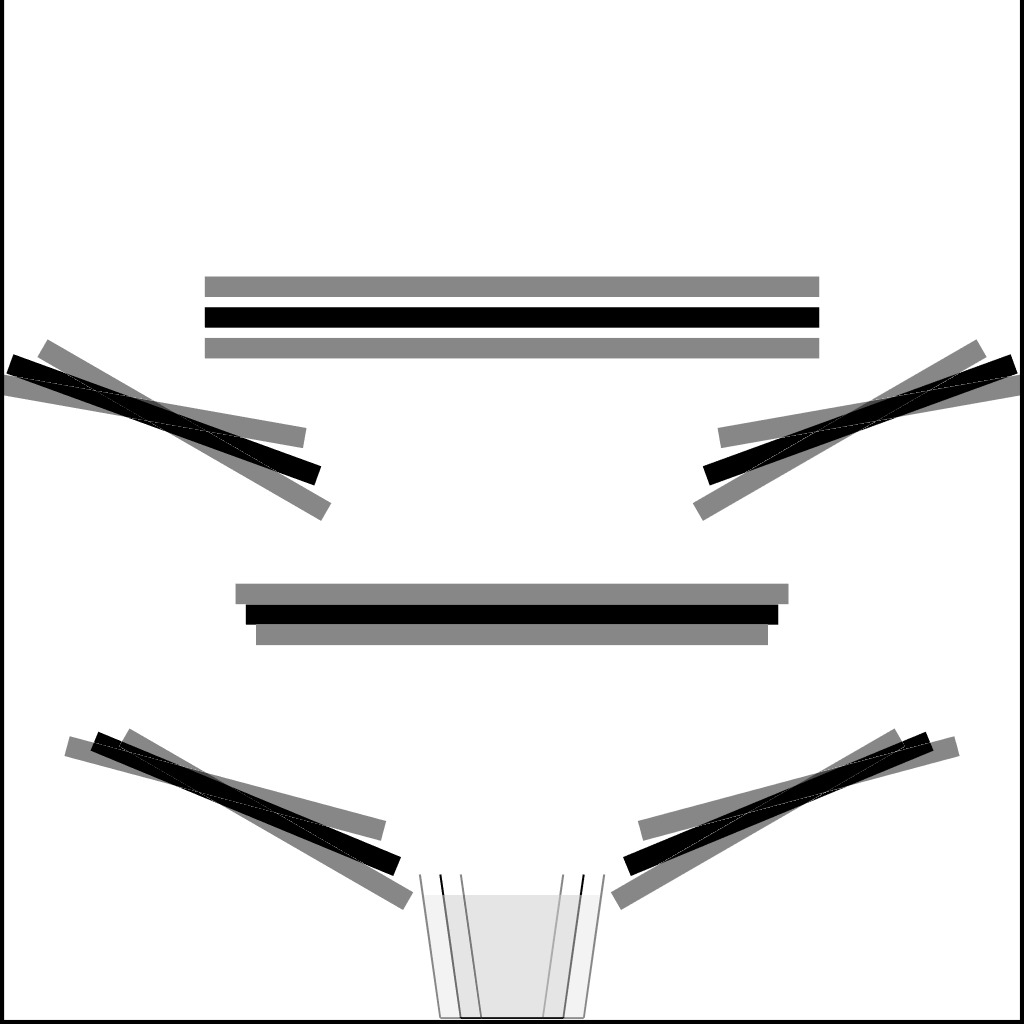} &
     \includegraphics[width=0.05\linewidth]{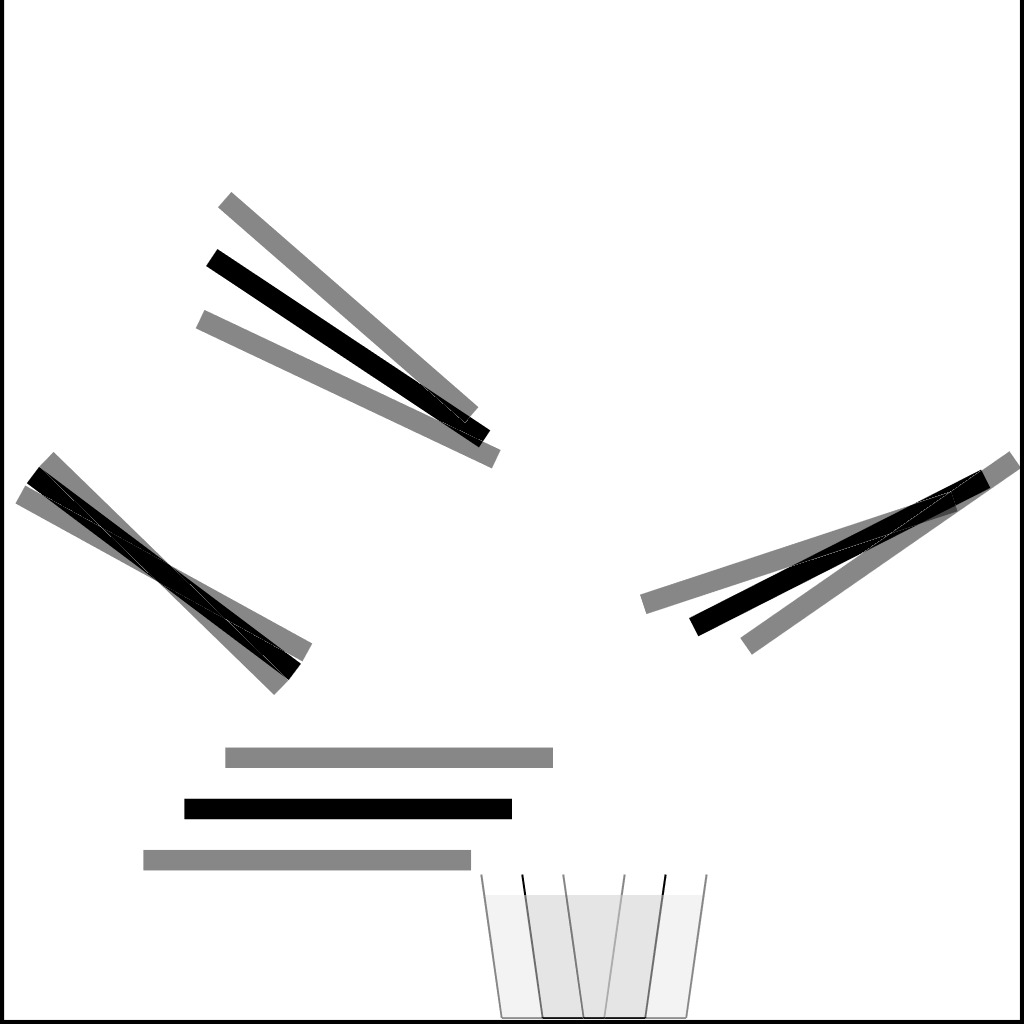} &
     \includegraphics[width=0.05\linewidth]{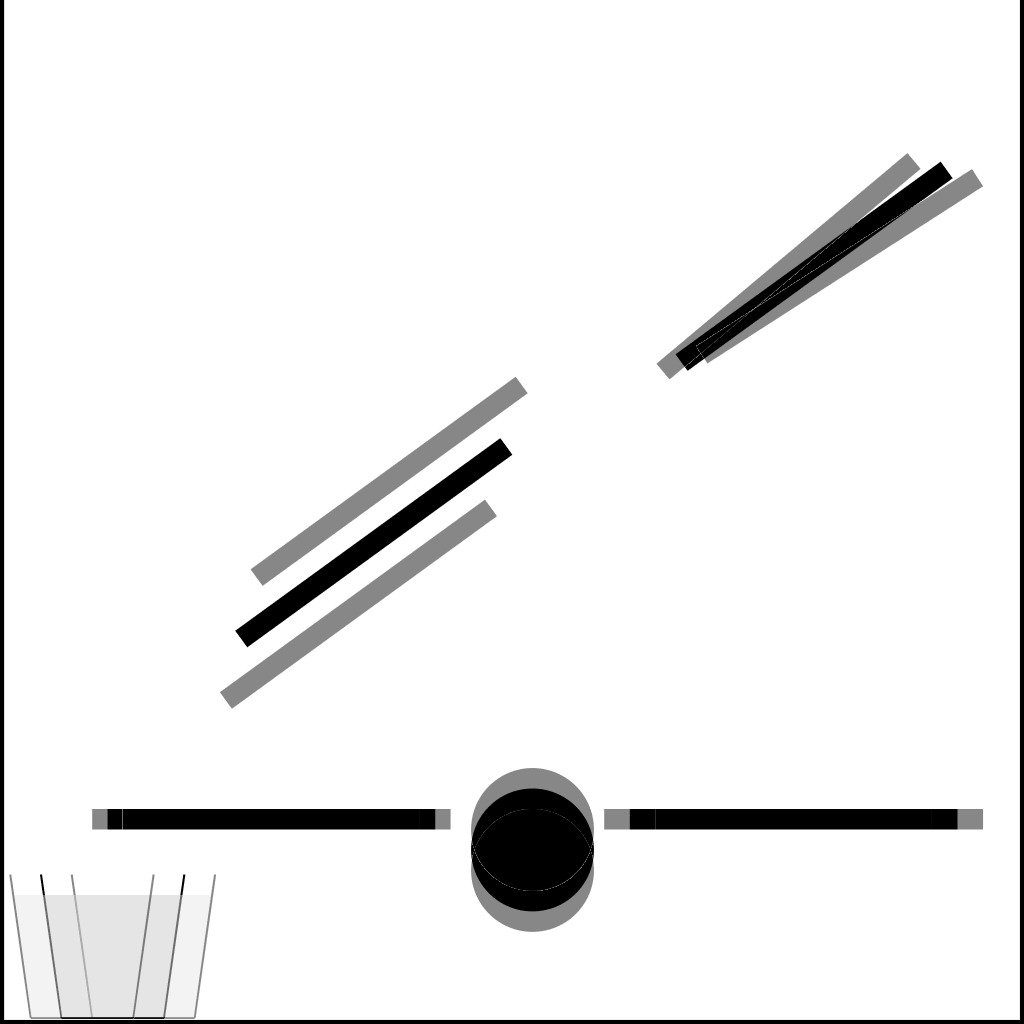} &
     \includegraphics[width=0.05\linewidth]{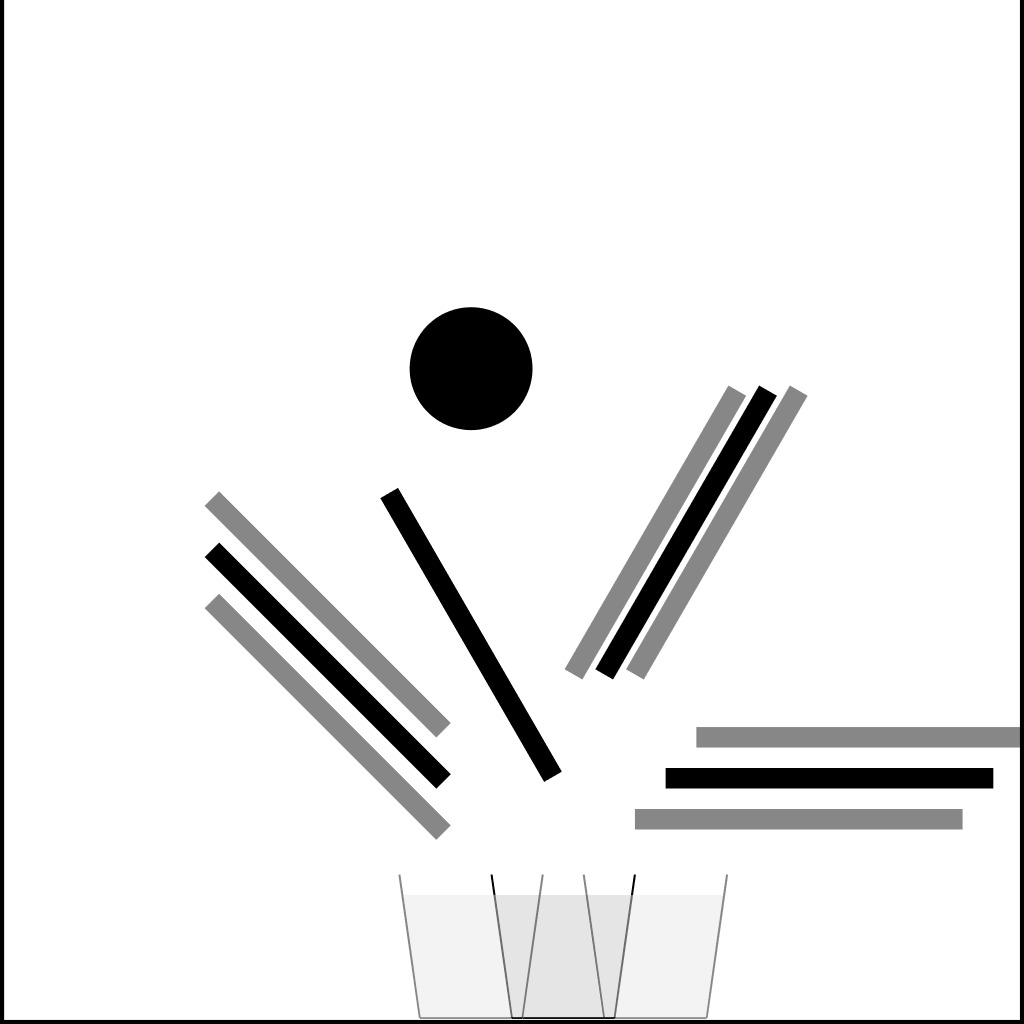} &
     \includegraphics[width=0.05\linewidth]{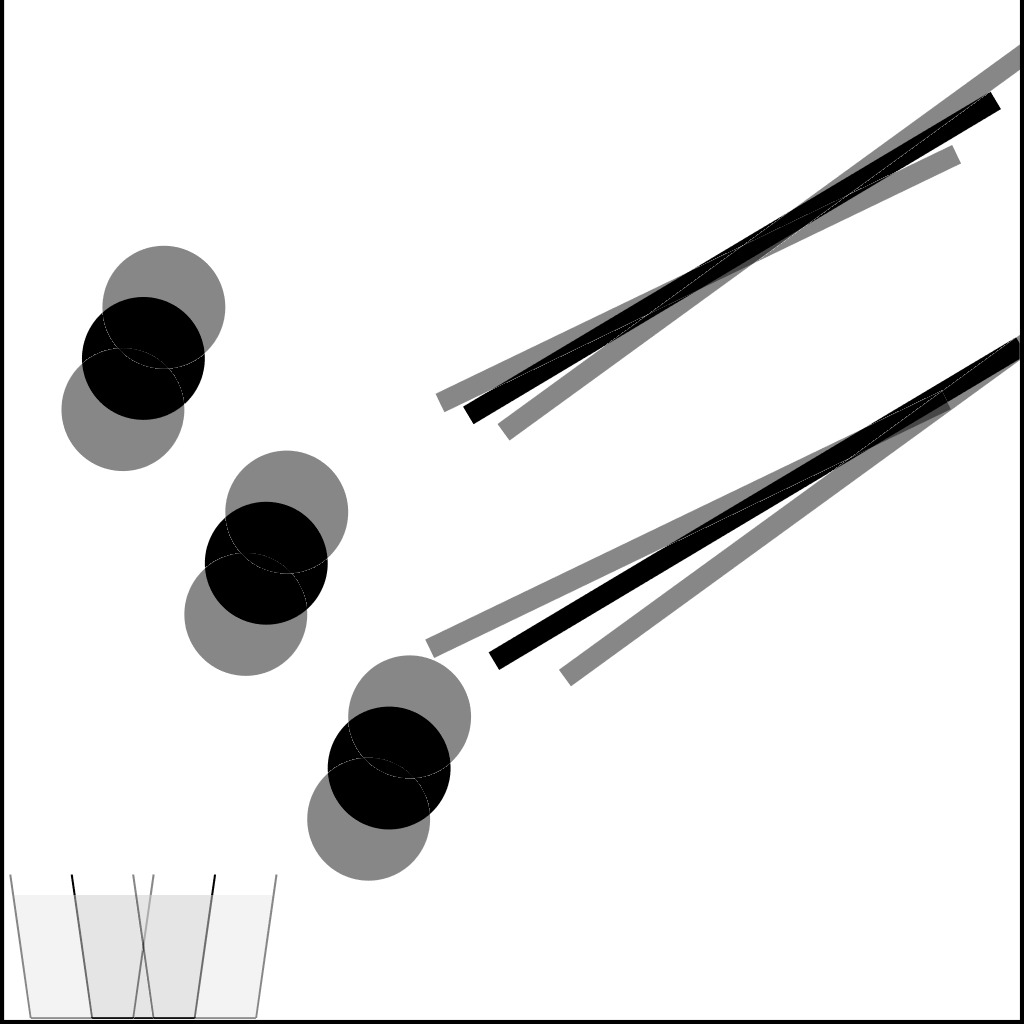} &
     \includegraphics[width=0.05\linewidth]{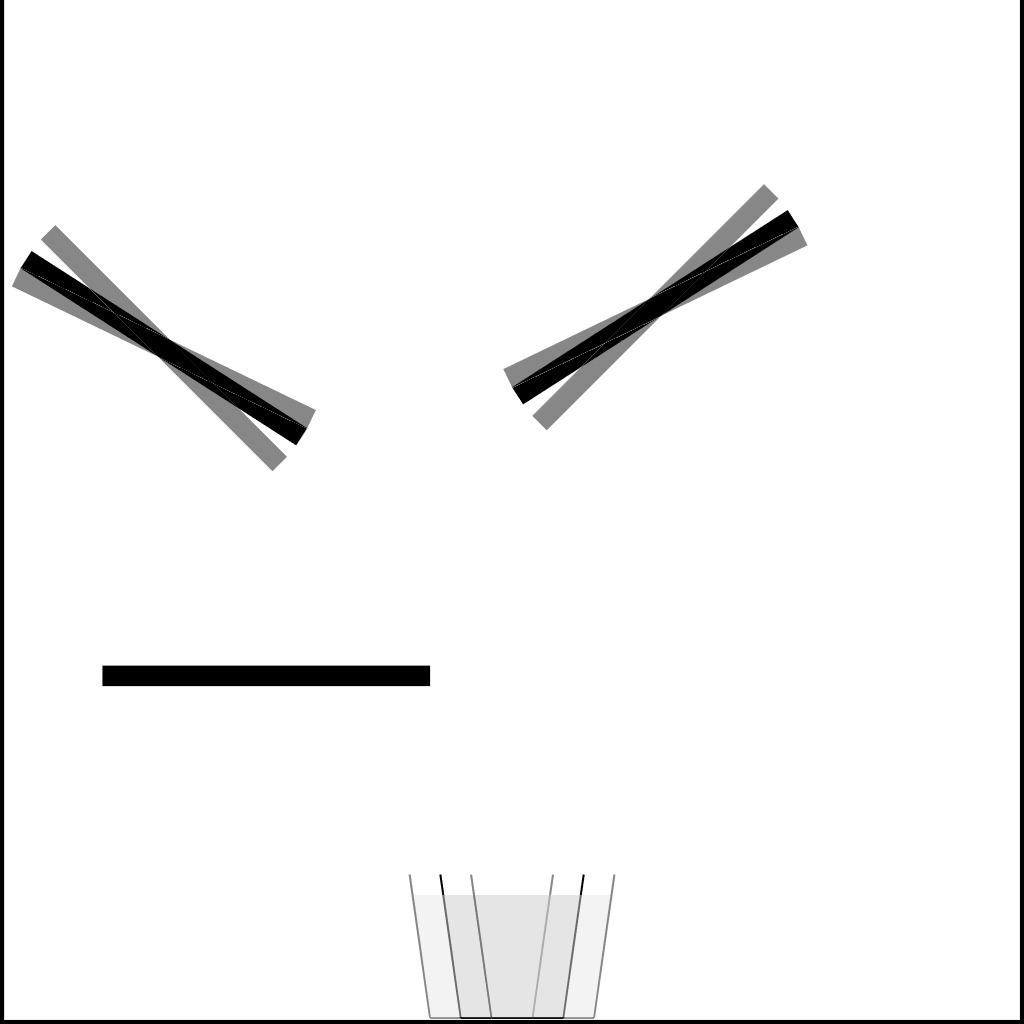} &
     \includegraphics[width=0.05\linewidth]{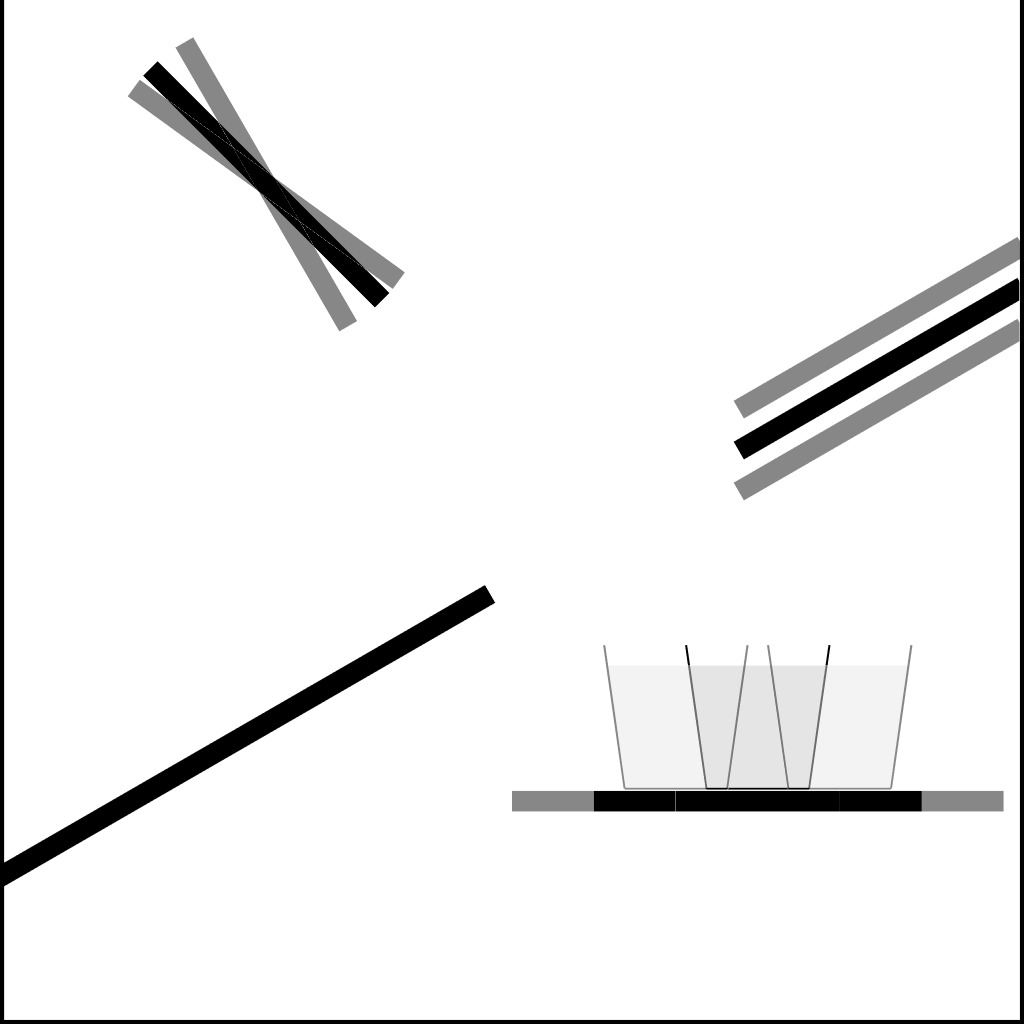} &
     \includegraphics[width=0.05\linewidth]{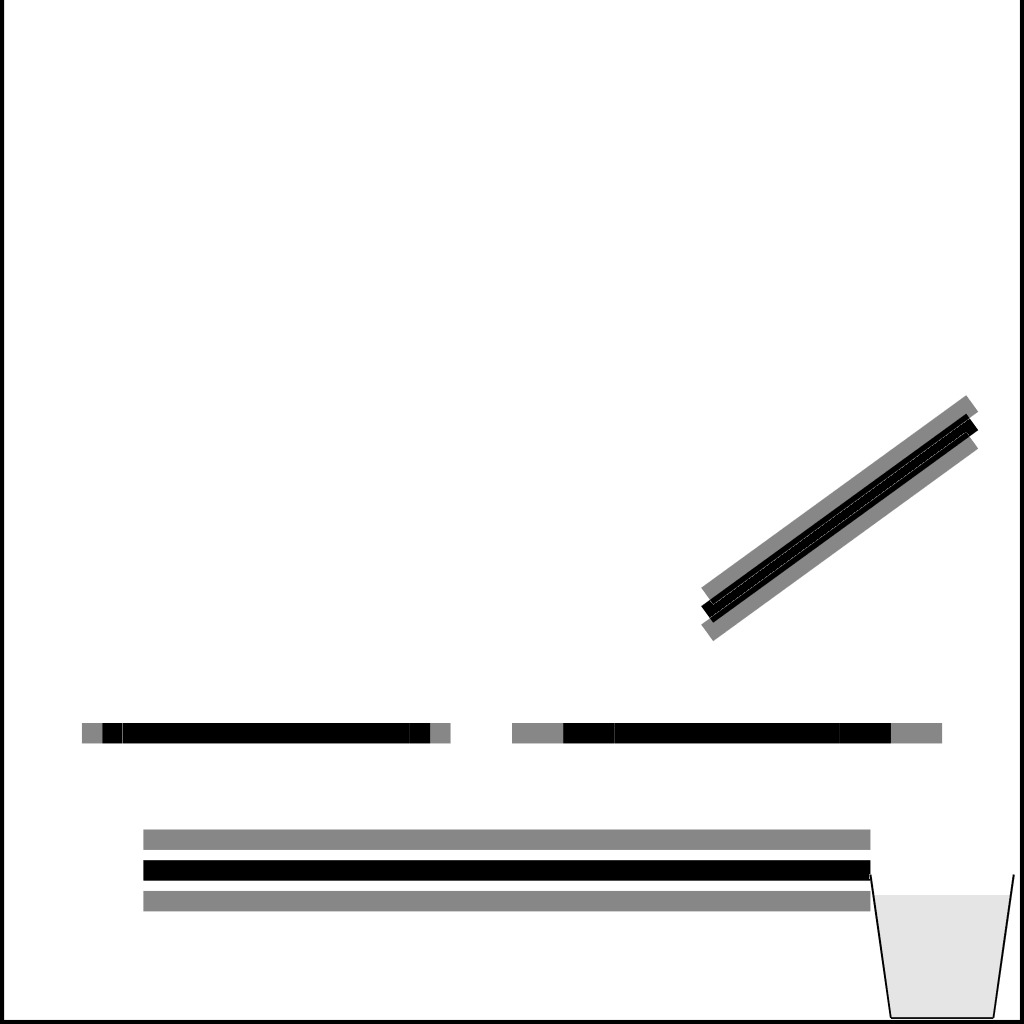} &
     \includegraphics[width=0.05\linewidth]{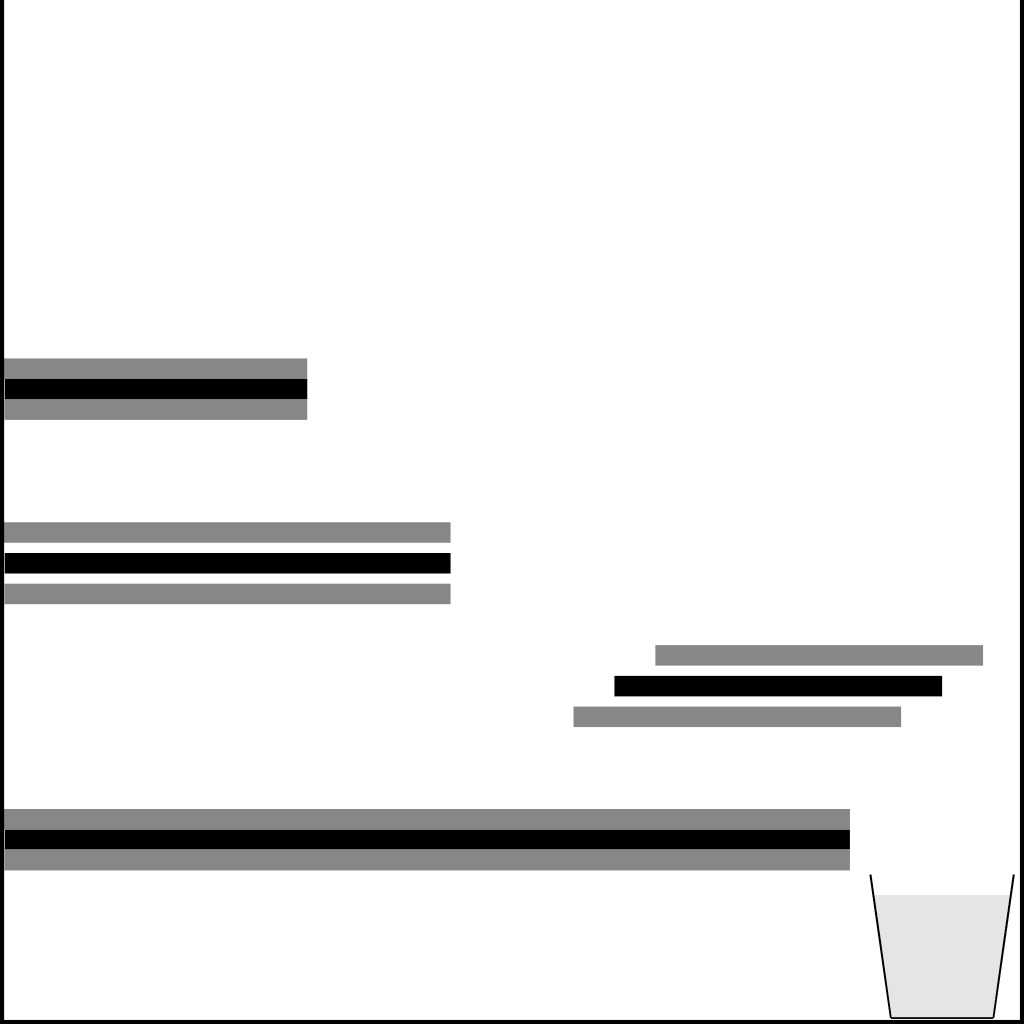} &
     \includegraphics[width=0.05\linewidth]{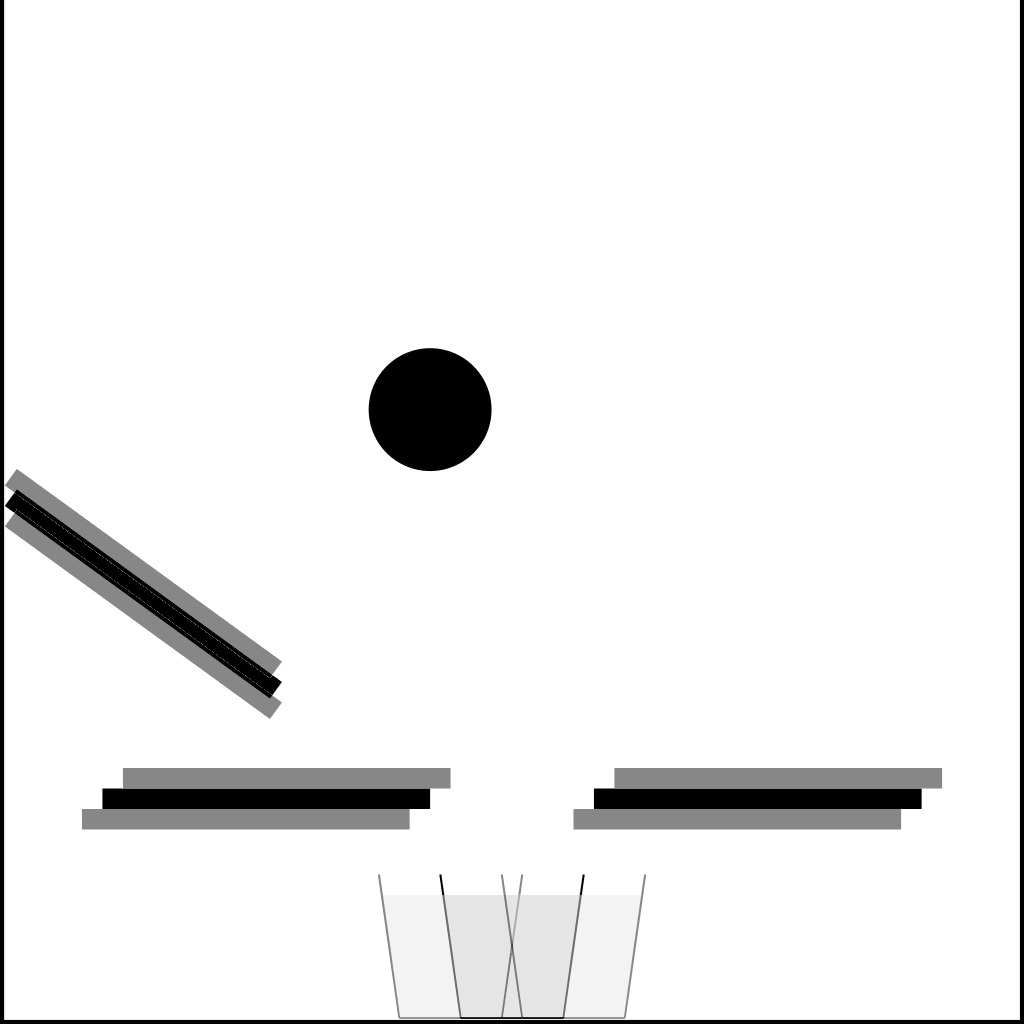} &
     \includegraphics[width=0.05\linewidth]{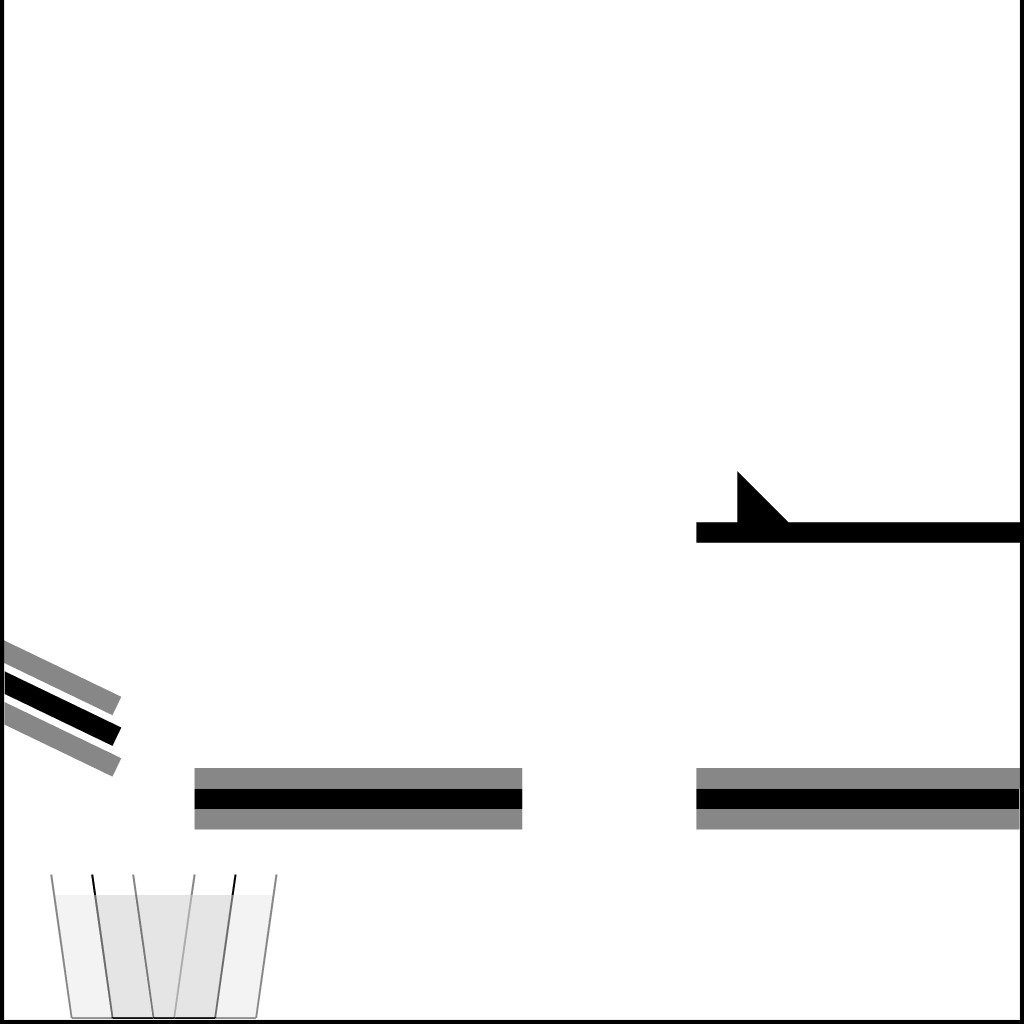} &
     \includegraphics[width=0.05\linewidth]{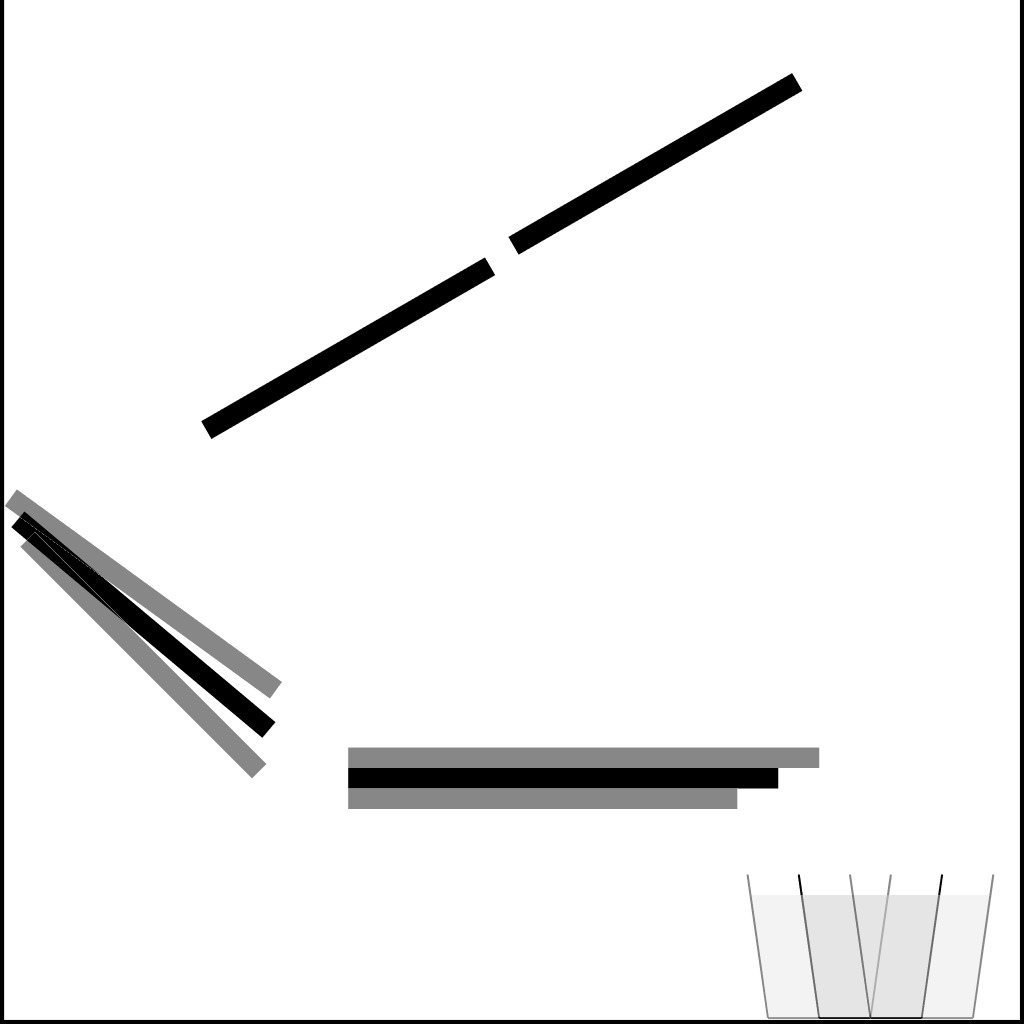}\\
     \midrule
     \multirow{2}{*}{Text-only} & LSTM & 43.52 & 41.17 & 45.39 & 40.36 & 46.12 & 42.03 & 39.29 & 34.90 & 44.47 & 44.91 & 47.95 & 50.14 & 44.20 & 52.50 & 39.92 & 46.15 & 43.06 & 45.28 & 48.47 & 42.84 \\
     & BERT  & 46.58 & 39.78 & 45.76 & 38.32 & 46.12 & 42.78 & 38.35 & 35.91 & 45.29 & 41.51 & 41.79 & 45.07 & 39.11 & 48.50 & 40.12 & 43.08 & 40.51 & 43.44 & 45.86 & 42.84 \\
     \midrule
     & LSTM-CNN-F & 51.34 & 48.69 & 51.22 & 40.14 & 52.71 & 47.59 & 47.37 & 37.25 & 46.72 & 46.29 & 57.69 & 48.22 & 44.20 & 53.62 & 43.99 & 52.75 & 50.00 & 50.21 & 53.22 & 52.69 \\
     Single & LSTM-CNN-L & 48.41 & 51.46 & 45.20 & 41.27 & 59.30 & 49.87 & 50.00 & 32.21 & 52.46 & 47.17 & 54.10 & 47.95 & 45.89 & 53.95 & 44.20 & 45.71 & 40.97 & 47.53 & 49.39 & 54.63 \\
     Frame & MAC-F & 47.43 & 48.39 & 50.28 & 41.72 & 50.78 & 50.89 & 47.93 & 44.63 & 48.57 & 45.28 & 51.28 & 47.67 & 40.94 & 53.95 & 45.62 & 50.77 & 45.83 & 47.95 & 50.92 & 50.90 \\
     & MAC-L & 44.99 & 45.31 & 50.85 & 42.63 & 56.59 & 51.39 & 51.13 & 43.96 & 46.72 & 41.38 & 52.05 & 45.21 & 42.11 & 53.73 & 47.45 & 47.69 & 48.61 & 48.52 & 51.69 & 51.34 \\
     \midrule
     & LSTM-CNN-V & 49.51 & 54.38 & 55.74 & 45.12 & 63.57 & 49.62 & 52.82 & 40.60 & 50.82 & 51.82 & 55.90 & 57.12 & 48.89 & 60.29 & 52.75 & 54.51 & 55.79 & 47.81 & 52.15 & 58.06 \\
     & MAC-V & 48.41 & 45.93 & 54.43 & 42.18 & 56.59 & 44.56 & 48.12 & 36.58 & 50.82 & 48.43 & 52.56 & 51.10 & 48.50 & 52.17 & 49.29 & 52.53 & 57.87 & 49.65 & 48.16 & 54.48 \\
     Video & TVQA & 44.38 & 47.16 & 42.00 & 38.10 & 46.12 & 41.77 & 43.98 & 30.20 & 45.08 & 44.03 & 44.10 & 48.36 & 42.89 & 56.84 & 41.34 & 44.62 & 40.05 & 43.86 & 46.17 & 45.22 \\
     & TVQA+ & 48.41 & 51.77 & 48.78 & 37.87 & 45.74 & 44.81 & 52.26 & 34.23 & 48.36 & 45.53 & 47.44 & 49.86 & 46.02 & 53.17 & 46.84 & 50.33 & 45.14 & 44.85 & 50.61 & 55.52 \\
     & G-SWM & 47.56 & 40.55 & 46.70 & 37.64 & 44.96 & 44.05 & 41.73 & 33.22 & 46.11 & 41.76 & 45.64 & 50.68 & 45.24 & 48.61 & 42.16 & 43.74 & 43.52 & 45.28 & 46.32 & 46.57 \\
     \midrule
     \multirow{2}{*}{Oracle} & LSTM-D & 58.92 & 51.15 & 61.96 & 59.86 & 67.83 & 65.82 & 54.89 & 61.41 & 63.52 & 58.99 & 66.15 & 61.64 & 54.11 & 60.73 & 62.32 & 60.88 & 61.81 & 56.28 & 54.91 & 61.94 \\
     & BERT-D & 83.62 & 79.72 & 89.27 & 88.89 & 96.12 & 86.58 & 84.77 & 92.62 & 81.15 & 85.28 & 88.72 & 94.52 & 82.40 & 82.65 & 91.04 & 85.27 & 88.89 & 85.05 & 86.04 & 85.67 \\
     \midrule
     & Human & 76.71 & 30.77 & 95.00 & 80.43 & 96.30 & 85.71 & 86.36 & 77.59 & 75.34 & 62.50 & 61.54 & 61.11 & 88.14 & 67.05 & 85.71 & 72.46 & 56.25 & 77.88 & 76.92 & 91.11 \\
    \bottomrule
    \end{tabular}
    }
\end{table*}

\renewcommand{\arraystretch}{1.3}
\begin{table}[!ht]
    \caption{Performances of the tested models per scene on the test set of the hard split of CRAFT.}
    \label{table:per-scene-hard-split}
    \centering
    \resizebox{0.5\linewidth}{!}{
    \begin{tabular}{clcccc}
    \toprule
     & \backslashbox{\bf{Model}}{\bf{Scene}} & \includegraphics[width=0.05\linewidth]{11.png} & \includegraphics[width=0.05\linewidth]{13.png}  & \includegraphics[width=0.05\linewidth]{17.png}  & \includegraphics[width=0.05\linewidth]{18.png}  \\
    \midrule
     \multirow{2}{*}{Text-only} & LSTM & 45.09 & 43.34 & 45.47 & 44.96 \\
     & BERT & 43.82 & 41.84 & 42.28 & 42.76 \\
     \midrule
     & LSTM-CNN-F & 43.23 & 32.77 & 46.76 & 44.48 \\
     Single & LSTM-CNN-L & 45.48 & 43.15 & 45.38 & 45.53 \\
     Frame & MAC-F & 50.66 & 44.24 & 45.99 & 47.34 \\
     & MAC-L & 47.83 & 44.00 & 47.56 & 46.39 \\
     \midrule
     & LSTM-CNN-V & 44.11 & 47.41 & 49.25 & 44.67 \\
     & MAC-V & 45.92 & 45.40 & 52.6 & 46.55 \\
     Video & TVQA & 44.70 & 42.91 & 43.05 & 43.72 \\
     & TVQA+ & 39.37 & 43.01 & 50.42 & 47.44 \\
     & G-SWM & 40.99 & 43.1 & 41.8 & 43.14 \\
     \midrule
     \multirow{2}{*}{Oracle} & LSTM-D & 67.32 & 54.82 & 56.91 & 55.62 \\
     & BERT-D & 87.59 & 83.40 & 85.73 & 84.47 \\
    \midrule
     & Human & 61.54 & 88.14 & 56.25 & 77.88 \\
    \bottomrule
    \end{tabular}
    }
\end{table}

\renewcommand{\arraystretch}{1.3}
\begin{table}[!ht]
    \caption{Performances of the tested models per question type on the test set of the easy split of CRAFT.}
    \label{table:question-type-easy-split}
    \centering
    \resizebox{\textwidth}{!}{
    \begin{tabular}{clcccccccccccc}
    \toprule
    & \bf{Model} & \bf{C/A} & \bf{C/N} & \bf{CF/N} & \bf{CF/O} & \bf{D/2Qs} & \bf{D/C} & \bf{D/C-T} & \bf{D/N-T} & \bf{D/N-V} & \bf{D/S} & \bf{D/TO} & \bf{All} \\
    \midrule
    \multirow{2}{*}{Text-only} & LSTM & 54.92 & 49.81 & 30.51 & 56.68 & 37.02 & 14.16 & 51.48 & 33.66 & 31.30 & 34.52 & 53.48 & 44.69 \\
    & BERT & 46.96 & 50.95 & 32.84 & 53.36 & 27.34 & 13.62 & 48.89 & 34.15 & 32.22 & 37.50 & 55.08 & 42.90 \\
    \midrule
    & LSTM-CNN-F & 54.14 & 51.34 & 36.02 & 58.24 & 30.80 & 31.98 & 54.53 & 35.12 & 31.30 & 46.68 & 52.58 & 49.07 \\
    Single & LSTM-CNN-L & 55.80 & 53.24 & 37.29 & 58.50 & 31.14 & 28.79 & 52.64 & 38.05 & 29.63 & 44.64 & 52.58 & 48.42 \\
    Frame & MAC-F & 54.03 & 51.72 & 36.23 & 55.49 & 35.99 & 32.76 & 52.84 & 35.12 & 31.11 & 44.98 & 53.83 & 48.10 \\
    & MAC-L & 50.61 & 48.85 & 37.08 & 55.59 & 32.53 & 35.10 & 53.05 & 38.54 & 30.74 & 43.28 & 53.65 & 47.83 \\
    \midrule
    & LSTM-CNN-V & 53.81 & 56.11 & 43.43 & 64.24 & 34.95 & 17.20 & 68.95 & 55.12 & 42.96 & 42.01 & 50.80 & 53.01 \\
    & MAC-V & 54.81 & 52.48 & 43.22 & 59.99 & 33.22 & 16.19 & 63.22 & 53.17 & 37.22 & 36.56 & 54.72 & 49.74 \\
    Video & TVQA & 54.81 & 51.72 & 33.26 & 59.07 & 29.07 & 11.75 & 50.54 & 37.56 & 30.19 & 33.76 & 52.23 & 44.71 \\
    & TVQA+ & 57.02 & 51.15 & 42.58 & 62.74 & 27.68 & 11.83 & 55.85 & 44.39 & 38.33 & 35.46 & 54.37 & 48.11 \\
    & G-SWM & 54.25 & 52.29 & 29.66 & 59.30 & 32.53 &  8.56 & 53.13 & 36.59 & 29.44 & 34.44 & 47.95 & 44.69 \\
    \midrule
    \multirow{2}{*}{Oracle} & LSTM-D & 52.82 & 49.81 & 41.74 & 58.10 & 31.83 & 68.09 & 68.37 & 41.46 & 41.11 & 73.72 & 53.12 & 59.53 \\
    & BERT-D & 70.28 & 65.27 & 69.07 & 81.77 & 46.37 & 96.42 & 97.90 & 72.20 & 85.56 & 98.21 & 96.61 & 86.20 \\
    \midrule
    & Human & 78.22 & 57.78 & 78.57 & 77.65 & 60.00 & 87.04 & 83.93 & 91.67 & 93.75 & 96.30 & 100.00 & 76.60 \\
    \bottomrule
    \end{tabular}
    }
\end{table}

\renewcommand{\arraystretch}{1.3}
\begin{table}[!ht]
    \caption{Performances of the tested models per question type on the test set of the hard split of CRAFT.}
    \label{table:question-type-hard-split}
    \centering
    \resizebox{\textwidth}{!}{
    \begin{tabular}{clcccccccccccc}
    \toprule
    & \bf{Model} & \bf{C/A} & \bf{C/N} & \bf{CF/N} & \bf{CF/O} & \bf{D/2Qs} & \bf{D/C} & \bf{D/C-T} & \bf{D/N-T} & \bf{D/N-V} & \bf{D/S} & \bf{D/TO} & \bf{All} \\
    \midrule
    \multirow{2}{*}{Text-only} & LSTM & 53.81 & 50.54 & 25.73 & 60.45 & 41.61 & 11.68 & 51.27 & 29.74 & 26.88 & 32.18 & 53.80 & 44.52 \\
    & BERT & 48.93 & 49.82 & 28.16 & 55.43 & 34.67 & 11.75 & 49.36 & 24.57 & 26.68 & 36.28 & 49.86 & 42.52 \\
    \midrule
    
    & LSTM-CNN-F & 48.93 & 46.76 & 27.67 & 50.94 & 39.78 & 15.74 & 45.87 & 30.60 & 29.25 & 30.68 & 50.14 & 40.64 \\
    First & LSTM-CNN-L & 50.60 & 48.74 & 25.24 & 58.47 & 31.39 & 19.44 & 50.87 & 30.17 & 23.52 & 37.07 & 53.12 & 44.66 \\
    Frame & MAC-F & 53.81 & 48.92 & 28.16 & 57.00 & 40.88 & 34.15 & 48.73 & 29.74 & 27.47 & 38.86 & 54.76 & 46.55 \\
    & MAC-L & 51.19 & 48.74 & 27.67 & 57.40 & 36.50 & 30.38 & 51.15 & 30.17 & 26.09 & 37.50 & 52.17 & 46.05 \\
    \midrule
    
    & LSTM-CNN-V & 52.86 & 50.36 & 32.77 & 57.94 & 42.70 & 14.56 & 61.29 & 28.88 & 29.45 & 33.55 & 48.91 & 46.50 \\
    & MAC-V & 51.43 & 50.90 & 35.19 & 57.40 & 47.81 & 16.33 & 62.24 & 37.07 & 31.62 & 33.26 & 52.04 & 47.31 \\
    Video & TVQA & 53.57 & 47.12 & 27.18 & 58.98 & 32.12 & 12.79 & 50.28 & 24.14 & 25.69 & 32.54 & 51.63 & 43.46 \\
    & TVQA+ & 53.10 & 47.84 & 29.61 & 58.64 & 25.91 & 13.90 & 58.23 & 27.16 & 24.90 & 31.82 & 52.17 & 45.12 \\
    & G-SWM & 50.60 & 51.62 & 31.07 & 51.11 & 37.59 & 12.86 & 50.72 & 25.86 & 26.48 & 36.78 & 52.72 & 42.47 \\
    \midrule
    
    \multirow{2}{*}{Oracle} & LSTM-D & 51.31 & 52.88 & 37.62 & 58.54 & 44.16 & 63.49 & 64.27 & 31.90 & 34.58 & 67.82 & 52.31 & 57.64 \\
    & BERT-D & 68.93 & 62.41 & 52.18 & 83.09 & 49.27 & 98.37 & 96.10 & 53.88 & 67.00 & 97.77 & 93.75 & 84.90 \\
    \midrule
    & Human & 78.22 & 57.78 & 78.57 & 77.65 & 60.00 & 87.04 & 83.93 & 91.67 & 93.75 & 96.30 & 100.00 & 76.60 \\
    \bottomrule
    \end{tabular}
    }
\end{table}

\clearpage
\subsection{Additional Examples}
\label{app_extraexamples}

\begin{figure*}[!ht]
\centering
\includegraphics[width=\linewidth]{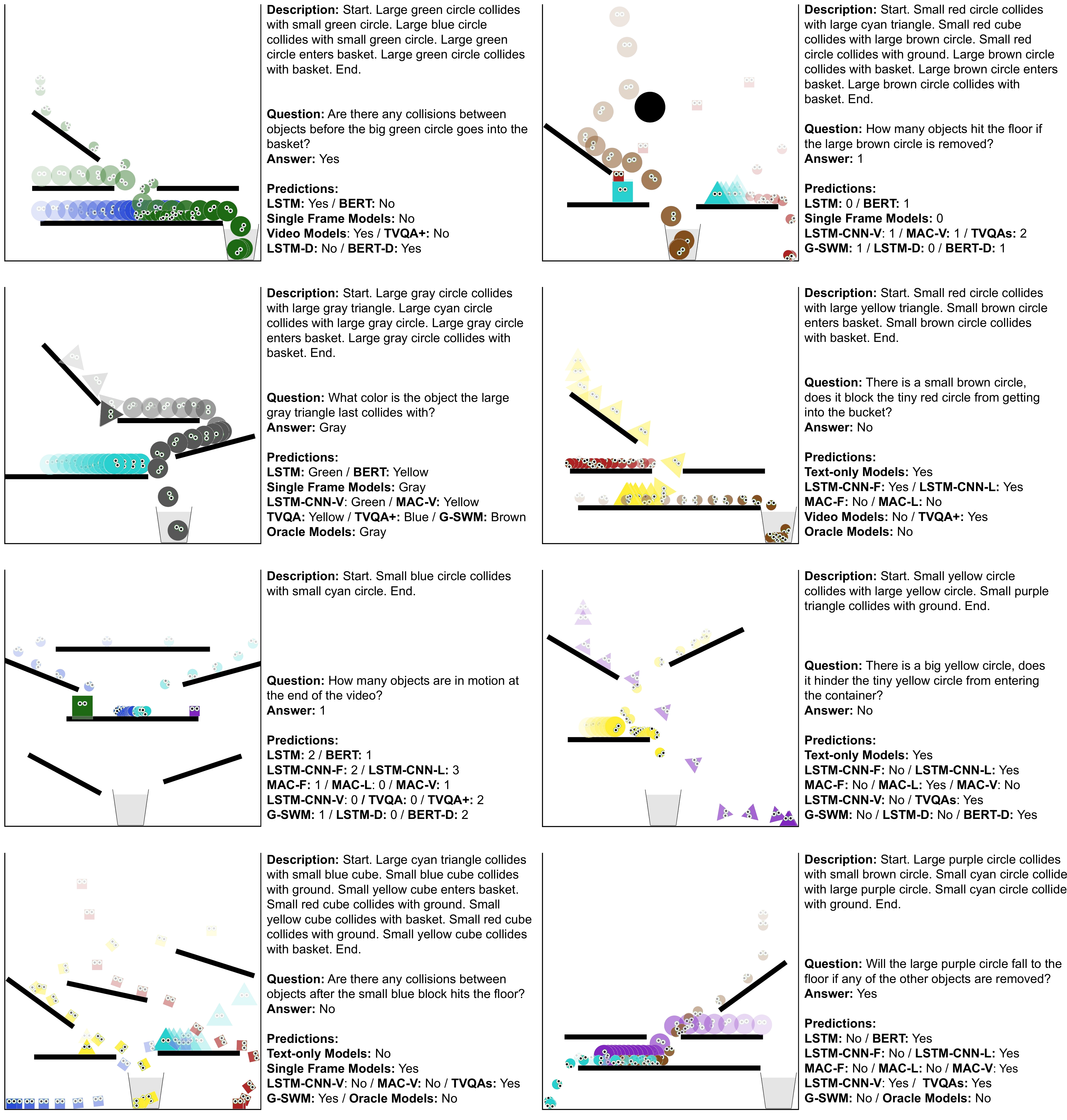}
\caption{Example model predictions. The examples on the left belong to the descriptive category and the right column contains examples from the other categories.}
\label{qualitative_results}
\end{figure*}

Figure~\ref{qualitative_results} provide some additional sample CRAFT questions together with the oracle descriptions and the baseline model predictions.

\subsection{Human Evaluation}
\label{human_evaluation}
The data from human participants were collected online via Qualtrics. The approximate time to complete the study was between 20 and 30 minutes. Participants did not take any bonus or wage. They attended the study voluntarily. The personal identifying information was not obtained. There were not an expected negative outcomes of the study on participants, but they could leave the study whenever they want. 
Koç University's Institutional Review Board approved the study (Protocol no: 2021.164.IRB3.073).

\begin{figure}[!h]
\centering
\includegraphics[width=0.9\linewidth]{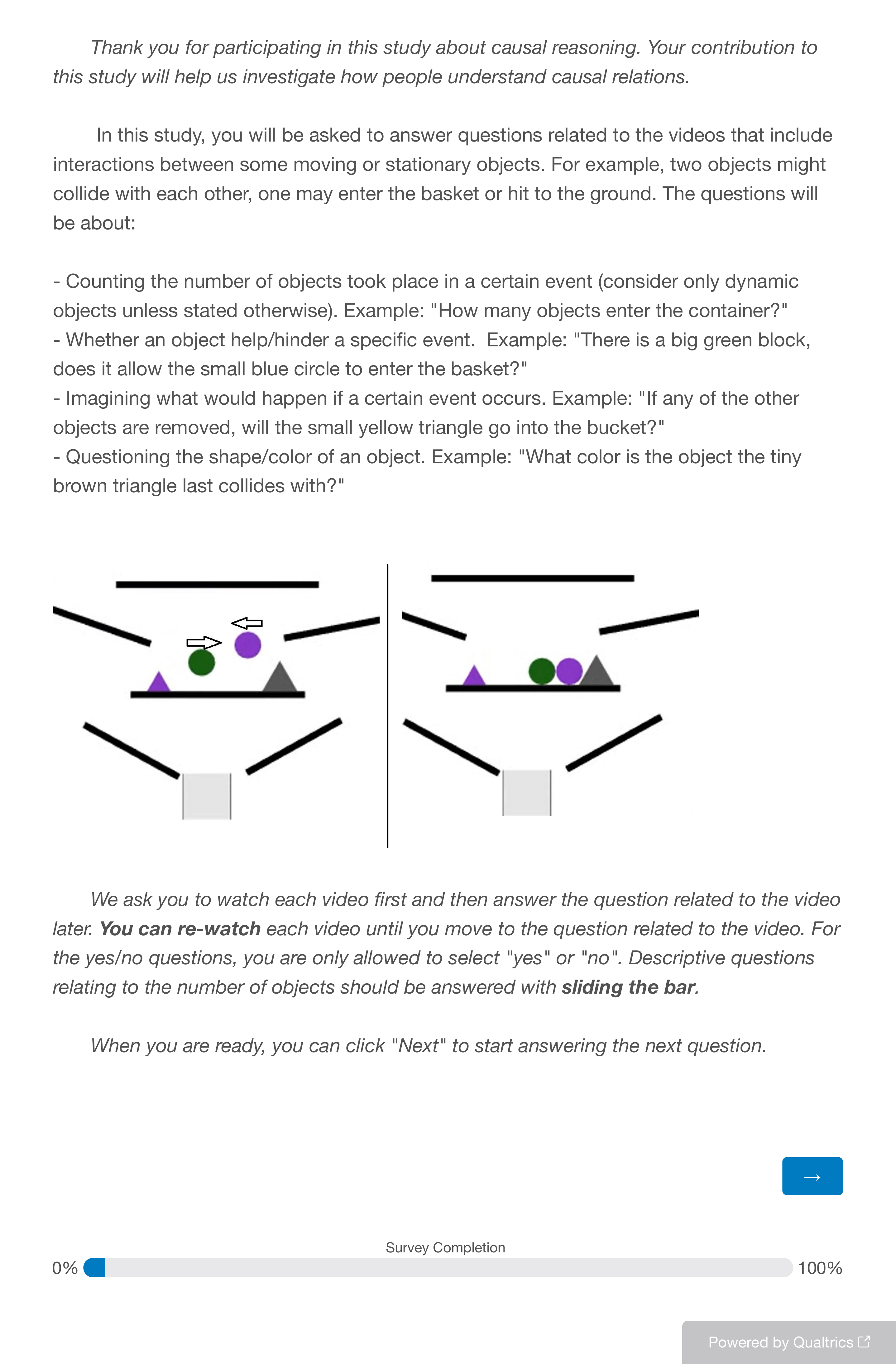}
\caption{The information form of the human evaluation study.}
\label{human_info_form}
\end{figure}

For the human evaluation, the participants saw the videos and multiple choice questions. The instruction page that was given to participants is shown in Figure ~\ref{human_info_form}.

\clearpage
\twocolumn
\section{Datasheet for CRAFT}
\label{datasheet}
This document is prepared in accordance with the guideline suggested in Datasheets for Datasets \citep{gebruDatasheetsDatasets2020}, the most updated version can be found \href{https://arxiv.org/abs/1803.09010}{\underline{here}}.\\
\begin{mdframed}
\vspace{-0.25cm}\subsection*{Motivation}
\end{mdframed}

\noindent\textbf{For what purpose was the dataset created?}

\noindent CRAFT was created in order to facilitate research on understanding and closing the gap between the capabilities of human intelligence and artificial systems in grasping and reasoning about physical relationships between different objects in an environment through vision and language.\\

\noindent\textbf{Who created this dataset (e.g., which team, research group) and on behalf of which entity (e.g., company, institution, organization)?}

\noindent The dataset was created by Tayfun Ates, M. Samil Atesoglu, Cagatay Yigit, Erkut Erdem from Hacettepe University and Ilker Kesen, Mert Kobas, Aykut Erdem, Tilbe Goksun and Deniz Yuret from Ko\c{c} University.\\

\noindent\textbf{Who funded the creation of the dataset?}

\noindent CRAFT was supported in part by GEBIP 2018 Award of the Turkish Academy of Sciences to E. Erdem and T. Goksun, BAGEP 2021 Award of the Science Academy to A. Erdem, and AI Fellowship to Ilker Kesen provided by the KUIS AI Center.\\

\begin{mdframed}
\vspace{-0.25cm}\subsection*{Composition}
\end{mdframed}
\noindent\textbf{What do the instances that comprise the dataset represent (e.g., documents, photos, people, countries)?}

\noindent The instances of CRAFT include a video, a question about the video, its answer, the functional program which is the ground-truth process that is used to answer the question, the states of dynamic objects and static scene elements at the start of the simulation and at the end of the simulations, causal graph of the events occurred in the video, variation videos which are created removing each dynamic object one by one, and lastly the states of objects and causal graphs for variation videos.\\

\noindent \textbf{How many instances are there in total (of each type, if appropriate)?}

\noindent CRAFT contains 58K video and question pairs that are generated from 10K videos from 20 different virtual environments.\\

\noindent \textbf{Does the dataset contain all possible instances or is it a sample (not necessarily random) of instances from a larger set?}

\noindent Please refer to Section 3 of the main paper for a detailed description of the sampling procedure used to generate questions.\\
    
\noindent \textbf{What data does each instance consist of?}

\noindent The video and question-answer pairs are used as the basic components for this visual question answering study. The question about the video is asked to an artificial model or a human subject. The test containing multimodal inputs question the capabilities of the subject in understanding and reasoning about physical relationships occurring in an environment. We use other instances in the dataset to find answers to questions automatically and share them for further analysis if required. Functional programs can run on object states and causal graphs to find the answer. Moreover, they can be integrated in training process for different models as well. Similarly, if ground-truth information regarding object states and causal graphs can also be extracted. Furthermore, some questions require counterfactual analysis that we define using variation videos formally. In order to evaluate effect of an object on the scene, we remove it an re-simulate the environment. We share instances regarding variations for further analysis.\\

\noindent \textbf{Is there a label or target associated with each instance? If so, please provide a description.}

\noindent Each instance consists of a ground-truth answer associated with the question about a dynamic scene.\\

\noindent \textbf{Is any information missing from individual instances?}
\noindent We do not provide object-level segmentation maps.\\

\noindent \textbf{Are relationships between individual instances made explicit (e.g., users’ movie ratings, social network links)?}

\noindent Instances are generated from 20 different scene layouts with some randomization.\\

\noindent \textbf{Are there recommended data splits (e.g., training, development/validation, testing)?}

\noindent We share CRAFT with two different split alternatives that we call easy and hard settings. Both of the alternatives contain non-overlapping train, validation, and test set. There are 20 distinct layouts from which we created our virtual scenes for CRAFT. In easy setting, each split might contain images from all of the scene layouts. On the other hand, in hard setting, train, validation, and test splits contain images from 12, 4, and 4 of the 20 layouts, respectively. That is, in the hard setting, the corresponding test samples are generated from unseen scene layouts.\\

\noindent \textbf{Are there any errors, sources of noise, or redundancies in the dataset?}

\noindent The process that we followed to make sure that the answers are not affected much with the slight perturbations to the initial states is described in Section~3 of the main paper.\\

\noindent \textbf{Is the dataset self-contained, or does it link to or otherwise rely on external resources (e.g., websites, tweets, other datasets)?}

\noindent The dataset is self-contained.\\

\noindent \textbf{Does the dataset contain data that might be considered confidential (e.g., data that is protected by legal privilege or by doctor patient confidentiality, data that includes the content of individuals non-public communications)?}

\noindent No.\\

\noindent \textbf{Does the dataset contain data that, if viewed directly, might be offensive, insulting, threatening, or might otherwise cause anxiety?}

\noindent No.\\

\noindent \textbf{Does the dataset relate to people?}

\noindent No.\\

\noindent \textbf{Does the dataset identify any subpopulations (e.g., by age, gender)?}

\noindent No.\\

\noindent \textbf{Is it possible to identify individuals (i.e., one or more natural persons), either directly or indirectly (i.e., in combination with other data) from the dataset?}

\noindent No.\\

\noindent \textbf{Does the dataset contain data that might be considered sensitive in any way (e.g., data that reveals racial or ethnic origins, sexual orientations, religious beliefs, political opinions or union memberships, or locations; financial or health data; biometric or genetic data; forms of government identification, such as social security numbers; criminal history)?}

\noindent No.\\

\begin{mdframed}
\vspace{-0.25cm}\subsection*{Collection Process}
\end{mdframed}\textbf{How was the data associated with each instance acquired?} 

\noindent All instances of CRAFT are generated automatically using a physics engine.\\

\noindent \textbf{What mechanisms or procedures were used to collect the data (e.g., hardware apparatus or sensor, manual human curation, software program, software API)?}

\noindent We use Box2D physics simulator~\citep{catto2010} to create our visual scenes, extract object states and causal graphs. Furthermore, we extend the work CLEVR \citep{johnson2017clevr} to create CRAFT questions and answers.\\

\noindent \textbf{If the dataset is a sample from a larger set, what was the sampling strategy (e.g., deterministic, probabilistic with specific sampling probabilities)?}

\noindent The dataset is generated from scratch and it does not depend on an already existing dataset.\\

\noindent \textbf{Who was involved in the data collection process (e.g., students, crowdworkers, contractors) and how were they compensated (e.g., how much were crowdworkers paid)?}

\noindent Authors prepared the scripts which create visual and textual data automatically.\\

\noindent \textbf{Over what time-frame was the data collected?}

\noindent Data generation scripts ran about 51 hours to create 9917 videos and 57524 questions.\\

\noindent \textbf{Does the dataset contain all possible instances?}

\noindent Although we provide all instances for this version of CRAFT, it is possible for anyone to create new samples by running the scripts provided in our code repository.\\

\noindent \textbf{If the dataset is a sample, then what is the population?}

\noindent Please refer to Section 3 of the main paper for a detailed description of the sampling procedure used to generate questions.\\

\noindent It is possible the enlarge CRAFT by running existing scripts to obtain huge amount of data because of the randomness existing in video generation process as described in the paper. New dynamic objects, static scene elements, events can also be created to enrich CRAFT. Moreover, it is also possible to add new types of scene layouts and question categories or types. For example, CRAFT focuses on mostly physical reasoning. It is possible to add tasks questioning different capabilities of Humans such as spatial reasoning, planning, and so on. There is actually no limit for creating datasets similar to CRAFT.\\

\noindent \textbf{Were any ethical review processes conducted (e.g., by an institutional review board)?}

\noindent Ko\c{c} University’s Institutional Review Board approved the user study (Protocol No: 5152021.164.IRB3.073).\\

\noindent \textbf{Did you collect the data from the individuals in question directly, or obtain it via third parties or other sources (e.g., websites)?}

\noindent The data from human participants for the user study were collected online via Qualtrics.\\

\noindent \textbf{Were the individuals in question notified about the data collection?}
\noindent Yes.\\

\noindent \textbf{Did the individuals in question consent to the collection and use of their data?}
\noindent The participants of the user study are asked to sign a consent form.\\ 

\noindent \textbf{Has an analysis of the potential impact of the dataset and its use on data subjects (e.g., a data protection impact analysis)been conducted?}

\noindent Not applicable.\\

\begin{mdframed}
\vspace{-0.25cm}\subsection*{Preprocessing/Cleaning/Labeling}
\end{mdframed}

\noindent \textbf{Was any preprocessing/cleaning/labeling of the data done(e.g., discretization or bucketing, tokenization, part-of-speech tagging, SIFT feature extraction, removal of instances, processing of missing values)?}

\noindent There were two preprocessing steps applied to the dataset. Firstly, after creating a video and question-answer pair, we applied simple perturbations by changing certain values of dynamic objects slightly at the start of the simulation and re-simulated the video. If the answer to the question is changed in any of the variations, then we removed the video and the question pair from the dataset. Secondly, in order to obtain a dataset which is uniform as possible in all dimensions, we removed video and question pairs whose answers are dominant after the first perturbation filter.\\

\noindent By collecting this dataset, we had the chance to observe that although the artificial systems have demonstrated incredible progress in the past decade, there are still areas that should be investigated for them. Therefore, CRAFT can be considered as a sample dataset which will facilitate the research in closing the gap between humans and artificial systems.\\

\noindent Preprocessing steps achieve two main aims of ours. Firstly, we wanted to eliminate video and question pairs whose answers are inconsistent between different variations of the same video with small perturbations. We observed that these were the cases for which humans subjects had some troubles. Secondly, we wanted to make CRAFT difficult enough for machine reasoning models by aiming at avoiding learning shortcuts by selecting the most frequent answers in answering questions. The second step of preprocessing procedure mostly achieves this aim.\\

\noindent \textbf{Was the ``raw'' data saved in addition to the preprocessed/cleaned/labeled data (e.g., to support unanticipated future uses)?}

\noindent The raw data were saved, but were not made public.\\

\noindent \textbf{Is the software used to preprocess/clean/label the instances available?}

\noindent We plan to publicly release the software used to generate the scenes and the questions.\\

\begin{mdframed}
\vspace{-0.25cm}\subsection*{Distribution}
\end{mdframed}
\noindent \textbf{Has the dataset been used for any tasks already?}

\noindent We have used the dataset to train unimodal and multimodal baselines described in the paper. \\

\noindent \textbf{Is there a repository that links to any or all papers or systems that use the dataset?}

\noindent Links to the related papers will be listed in the project website at \url{https://sites.google.com/view/craft-benchmark}.\\

\noindent \textbf{What (other) tasks could the dataset be used for?}

\noindent Since the sample videos in our dataset include interactions between the objects themselves and the environment, they can be used in problems such as future state prediction and video generation.\\

\noindent \textbf{Is there anything about the composition of the dataset or the way it was collected and preprocessed/cleaned/labeled that might impact future uses?}

\noindent No.\\

\noindent \textbf{Are there tasks for which the dataset should not be used?}

\noindent No.\\

\begin{mdframed}
\vspace{-0.25cm}\subsection*{Uses}
\end{mdframed}
\noindent \textbf{Will the dataset be distributed to third parties outside of the entity (e.g., company, institution, organization) on behalf of which the dataset was created?}

\noindent CRAFT is publicly available at \url{http://github.com/hucvl/craft/}.\\

\noindent \textbf{How will the dataset will be distributed (e.g., tarball on website, API, GitHub)?}

\noindent The dataset is available through our project website and GitHub. Large dataset files are stored on Zenodo.\\

\noindent \textbf{When will the dataset be distributed?}

\noindent The dataset was first released in June 2021.

\noindent \textbf{What license (if any) is it distributed under?}

\noindent The dataset is released under \href{https://en.wikipedia.org/wiki/MIT_License}{MIT license}.\\

\begin{mdframed}
\vspace{-0.25cm}\subsection*{Maintenance}
\end{mdframed}
\noindent \textbf{Who is supporting/hosting/maintaining the dataset?}

\noindent CRAFT will be supported and maintained by the prime authors.\\

\noindent \textbf{Will the dataset be updated (e.g., to correct labeling errors, add new instances, delete instances)?}

\noindent Extending CRAFT in different directions is planned. All versions of CRAFT will be available at \url{http://github.com/hucvl/craft/}. 

\end{document}